# A Survey on Concept Factorization: From Shallow to Deep Representation Learning

Zhao Zhang[1,2,*], Yan Zhang[2], Mingliang Xu[3,*], Li Zhang[2], Yi Yang[4], and Shuicheng Yan[5]

[1] School of Computer Science and Information Engineering & Key Laboratory of Knowledge Engineering with Big Data (Ministry of Education), Hefei University of Technology, Hefei, China

[2] School of Computer Science and Technology (School of Software), Soochow University, Suzhou 215006, China

[3] School of Information Engineering, Zhengzhou University, Zhengzhou, China

[4] Centre for Artificial Intelligence, University of Technology Sydney, Sydney, NSW, Australia

[5] YITU Technology, China

*Corresponding author emails: cszzhang@gmail.com, iexumingliang@zzu.edu.cn

*Abstract*— The quality of obtained features by representation learning determines the performance of a learning algorithm and subsequent application tasks (e.g., high-dimensional data clustering). As an effective paradigm for learning representations, Concept Factorization (CF) has attracted a great deal of interests in the areas of machine learning and data mining for over a decade. Moreover, lots of effective CF-based methods have been proposed based on different perspectives and properties, but it still remains not easy to grasp the essential connections and figure out the underlying explanatory factors from current studies. In this paper, we therefore survey the recent advances on CF methodologies and the potential benchmarks by categorizing and summarizing current methods. Specifically, we first review the root CF method, and then explore the advancement of CF-based representation learning ranging from shallow to deep/multilayer cases. We also introduce the potential application areas of CF-based methods. Finally, we point out some future directions for studying the CF-based representation learning. Overall, this survey provides an insightful overview of both theoretical basis and current developments in the field of CF, which can also help the interested researchers to understand the current trends of CF and find the most appropriate CF techniques to deal with particular applications.

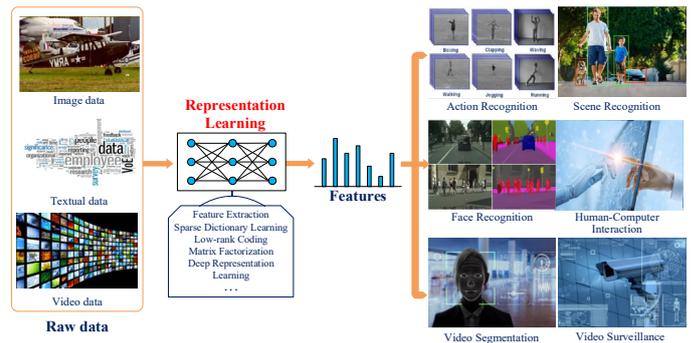

**Fig.1:** Application examples of representation learning (RL), where the pictures are from the Internet.

## I. INTRODUCTION

### A. Background

Learning compact features of high-dimensional data (e.g., image, document or video) via representation learning (RL) is a long-standing and challenging topic in the communities of data mining, pattern recognition, computer vision and neural networks [45]. To be more specific, RL algorithms play an important role in evaluating the performance of a learning algorithm for distinguishing and recognizing different objects [191-192][175], since RL methods can effectively simplify complex input data, eliminate invalid information and extract useful information (or features) from the observed inputs [45-55][177]. The extracted features by RL can then be applied in various application scenarios, such as action/face/scene recognition, image/video segmentation, video surveillance, human-computer interaction and so on, as shown in Fig.1. Classical RL methods include subspace learning for feature extraction (FE) [67-90], sparse dictionary learning (SDL) [56-66][181], robust low-rank coding (LRC) [91-104], matrix factorization (MF) [1-6][105-113][158] and deep RL (DRL) [35-41] [114-119], etc. Specifically, MF aims to factorize input data into the product of several matrices, of which Principal Component Analysis (PCA) [1], Singular Value Decomposition (SVD) [2], Independent Component Analysis (ICA) [3], Vector Quantization (VQ) [4], Nonnegative Matrix Factorization (NMF) [5] and Concept Factorization (CF) [6] are representative models. Compared with NMF and CF, existing PCA, SVD, ICA and VQ methods not only allow the existence of negative factors or subtractive combinations in representations, but also can only obtain compact features in a linear way. However, in reality negative values in the factorization matrices are mostly unexplained, such as the pixel value in image data and the word frequency in text data are all non-negative. To solve these issues, novel MF paradigms, i.e., NMF and CF, were proposed one after another. Given a data matrix, NMF decomposes it into a product of two nonnegative factors, which can obtain parts-based representations due to the nonnegative constraints. By this way, the interpretability of the factorization process can be enhanced, and moreover NMF can perform nonlinear dimensionality reduction for high-dimensional data. Although NMF has obtained great success for image processing and document clustering, it cannot directly be performed in the reproducing kernel Hilbert space (RKHS). As a variant of NMF, CF encodes each concept as a linear combination of data points and then represents each data point by a linear combination of the concepts, so it can be operated in any data representation space, including kernel space.

Starting from 2004, CF has been receiving much attention and achieving fast development in the fields of RL and high-dimensional data clustering in recent decade. In spite of raising many CF-based algorithms, however, to the best of our knowledge, there is still not a comprehensive survey to grasp the essential connections, figure out the underlying explanatory factors and categorize current developments on CF. Note that there already existed some survey/review papers on RL, such as the review on LRC [50], survey on SDL [51], review on sparse representation

(SR) [52], reviews on NMF and its extensions [53-54][184-185], and some other ones on RL from different perspectives [45-49][55]. In addition, there also existed some surveys on MF models, which are more related to this work. For example, NMF-based methods were reviewed in [53], which mainly discussed the theoretical relationship between NMF and clustering. It also makes a gentle introduction to how the clustering problem can be interpreted in a matrix factorization setting; another comprehensive review of NMF and its variants were proposed in [54], which mainly focused on the theoretical research into NMF. The principles, basic models, properties and algorithms of NMF models are also summarized systematically; the review in [184] mainly summarized the applications of NMF and its extensions in the recommendation systems, while [185] introduced the fundamentals and properties of NMF and divided the current NMF-based methods into basic ones and improved ones. Some open problems in the development of NMF has also been analyzed in [185]. It is clear that although these review or survey papers on MF also discussed the nonnegative factorization problem, their main focus and major goals are clearly different from this work that focuses on reviewing current CF-based models in a comprehensive way. Specifically, this present work mainly provides the detailed theoretic analysis on comparing different CF-based algorithms, concludes the advantages/disadvantages of existing CF-based methods, analyzes the relationship of the algorithms within each sub-category, analyzes which kinds of CF algorithms are useful for specific tasks or data distributions, evaluates the performance of different CF methods, discusses the possible application scenarios of CF algorithms, and points out some future directions, which will be meaningful, helpful and indispensable for the researchers and newcomers in the field.

*B. Contribution of this Survey*

In this work, we aim at presenting a comprehensive survey about current CF-based methods. Specifically, the major contributions of this survey paper are summarized as follows:

1) Almost all the existing CF methods have been reviewed, categorized and summarized in this paper to help newcomers to obtain a holistic understanding in related fields. Specifically, we mainly introduce the motivation, main idea, problem formulation, and advantage/disadvantage for each method;

2) The detailed analysis about the relationships and differences of various types of CF algorithms are provided. In this work, we first divide current CF-based algorithms roughly into shallow and deep ones in general, and then further divide shallow methods into unsupervised, semi-supervised and fully-supervised ones. Finer divisions are also given according to different characteristics and properties. To help the researchers, readers and newcomers thoroughly and easily understand, we clearly summarize the relationships of the methods in each category and the difference of methods in different categories;

3) The trend of development of CF-based methods has been analyzed. Specifically, we have shown the milestones of the CF-based methods and moreover also summarized the major development trends of the CF-based variants with the practical needs of emerging applications;

4) Extensive experimental results over real-world databases are reported in this survey to compare certain representative shallow and deep CF models. The detailed analysis about the experimental results is also provided;

5) We list the potential practical applications of CF-based methods, such as dimensionality reduction, high-dimensional data clustering, image processing, text/document processing, recommendation system, and so on. Based on the properties and distributions of different real data, we can potentially decide which kinds of CF algorithms will useful for specific tasks, e.g., we can choose the sparsity-constrained CF methods for processing document data due to the sparse properties of document data and select the locality-preserving CF methods for processing the data with manifold structure and distributions, which may guide the related researchers to apply specific CF models for different practical applications;

6) Although the field of CF-based representation learning has achieved rapid development, some remaining issues are still not solved. As such, we also point out certain future directions for relevant researchers, such as the problems on optimization, initialization, determination of rank of factorization, evaluation criteria, how to incorporate task-driven characteristic into CF, and how to design effective deep CF structures, etc.

*C. Content Organziation of this Survey*

In this survey, we first divide current CF-based algorithms into two major categories, i.e., CF-based single-layer and CF-based deep/multilayer methods. The single-layer methods are further divided into unsupervised, fully-supervised and semi-supervised ones depending on how much supervised prior knowledge can be used. To grasp the intrinsic relations and property, we further divide the unsupervised ones into five sub-categories:

1) **Locality-preserving CF.** For effective RL, retaining the locality manifold information of data in feature space is important, so locality-preserving CF is a major and the most popular strategy to improve CF, which has received much attention in recent years. Note that since different locality-preserving strategies can be used, we further divide this category into four parts based on the locality-preserving strategies, i.e., graph-regularized CF, local coordinate coding driven CF, self-representation based CF and other strategies, which are respectively described as follows:

✧ *Graph-regularized CF methods* incorporate the graph Laplacian regularization into the framework of CF. Therein, Locally Consistent CF (LCCF) [7] was the first method to use the graph Laplacian to smooth the representation and extract concepts with respect to the intrinsic manifold structures. Inspired by the idea of LCCF, some other representative variants have also been recently proposed, such as Dual-graph Regularized CF (GCF) [8], Adaptive Dual-Graph Regularized CF (ADGCF) [13], Graph-Regularized Local Coordinate CF (GRLCF) [9], Graph-regularized CF with Local Coordinate (LGCF$_1$) [10] and Multiple Graph Regularized CF with Adaptive Weights (MCFAW) [11]. Although the locality can be clearly retained by these graph-based CF methods, they still have one glaring flaw, i.e., it is tough to choose an optimal number of nearest neighbors to define the neighborhood graph. To overcome this issue, researchers proposed optimized adaptive-graph based CF methods recently, e.g., CF with Adaptive Neighbors (CFANs) [12], CF with Optimal Graph Learning (CF-OGL) [126], Graph-Regularized Local coordinate CF with CLR (GRLCF$_{CLR}$) [9] and Robust Flexible Auto-weighted Local-Coordinate CF (RFA-LCF) [14-15], etc. These optimized weighting strategies avoid the tricky issue of choosing the optimal number of neighbors in constructing the neighborhood graph effectively.

✧ *Local coordinate coding based CF methods* incorporate the idea of local coordinate coding (LCC) into CF to preserve local information in data. The first method of this category is Local Coordinate CF (LCF) [16-17] that incorporates the local coordinate coding as a locality constraint, so that it can exploit the sparsity and locality of samples at the same time. Subsequently, researchers have also incorporated the idea of LCC into the graph-based CF methods, for instance Graph-based Local concept coordinate factorization (GLCF) [18], GRLCF [9], LGCF$_1$ [10] and RFA-LCF [14-15].

✧ *Self-representation based CF methods* are mainly motivated by the success of exploiting the self-representation of data, in which input data are regarded as a dictionary, which provided new idea for deriving self-expressive CF. Specifically, CF is regarded as a nonnegative self-expression model with a learning-based dictionary to reveal the global structure of input data. Classical methods of this kind include Self-Representative Manifold CF (SRMCF) [20] and Joint Structured Graph Learning and clustering based CF (JSGCF) [21].

✧ Some other locality-preserving ways for local CF can also be used, which are different from above-mentioned ones. For example, Local regularization CF (LRCF) [30] retains local information by introducing the local learning regularization; Similarity-based CF (SCF) [127] and its robust version, Robust SCF (RSCF) [127], mainly rely on a similarity matrix; Sparse Dual Regularized CF (SDRCF) [120] obtains the locality-preserving representations with the help of SR [160-161]. Local Sensitive Dual Concept Learning (LSDCL) [183] adopts the local sensitive loss function to characterize the local structures of data. Another related method is called CF with Local Centroids (CFLCs) [186] that retains the local data manifold with the aid of multiple local centroids.

2) **Kernel CF.** To improve the clustering performance, Li *et al.* [19] first come up with a new method called Manifold Kernel CF (MKCF) [19] that incorporates the manifold kernel learning into CF. By the means of manifold kernel learning, the intrinsic structure of samples can be discovered in the warped RKHS. Note that GLCF [18] also applies the manifold kernel learning, which is also a classical method in this kind. Compared with MKCF and GLCF that are all single-kernel methods, both Globalized Multiple Kernel CF (GMKCF) [136] and Discriminative Multiple Kernels [182] have designed the multiple-kernel models to solve this tough issue, i.e., how to choose an optimal kernel function in real applications.

3) **Robust CF.** Traditional CF mainly aims at learning compact representation of original raw data, but real data usually have various noise and outliers, so their performance may be degraded by the negative effects of them. To deal with this issue, researchers have paid extensive efforts to the robust CF. For example, ADGCF [13] performs CF on the obtained features by feature selection, rather than original data. Robust and Discriminative CF (RDCF) [121] considers the noise as a sparse component of matrix factorization and makes it apart from raw data. In addition, RFA-LCF [14], Robust Local Learning and Discriminative CF (RLLDCF) [125], RSCF [127] and Correntropy-based Graph-regularized CF (GCCF) [124], are proposed to replace the noise-sensitive Frobenius norm using robust norms to encode the reconstruction error.

4) **Unsupervised Discriminative CF.** For the unsupervised CF, how to obtain the discriminant new representation is also a hot topic, as there is no supervised label information available. To learn discriminant representations in unsupervised scenario, RDCF [121], RLLDCF [125] and Structured Discriminative CF (SDCF$_2$) [128] are proposed, which are able to discover the intrinsic discriminant structure of data space without the supervised label information.

5) **Multi-view CF.** Multi-view learning is a hot topic in machine learning [144-146][162-163], because multi-view data can be observed in various real-world applications. However, most existing CF methods are based on single-view, which cannot handle complex multi-view data. As such, researchers also explored several multi-view CF methods, and representative methods include the Multi-view Clustering via CF (MVCC) [122], and Adaptive Structure CF for Multiview Clustering (MVCF) [123], which successfully extend traditional single-view based CF methods to multi-view scenarios.

Based on the above categories, the involved methods will potentially have abilities to learn features with different characteristics, which will be able to satisfy the specific requirements of various practical applications. Generally speaking, these unsupervised methods can be used to deal with specific tasks where label information of samples is unavailable, e.g., unsupervised single-view/multi-view clustering, robust feature learning, manifold feature extraction, and so on. To be specific, we can choose appropriate CF-based methods for specific tasks according to the properties and distributions of data in reality. For example, locality-preserving CF methods are suitable for handling samples with manifold distribution, kernelized CF methods will be suitable to deal with the linearly-inseparable problems by nonlinear mapping, robust CF methods will be suitable for handling the raw data with noisy information, discriminative CF methods can be used in the cases that higher discriminant performance is required for the learned features, and multi-view CF methods are usually required in the cases that training samples have different representations in multiple views or come from different sources.

It is noted that unsupervised CF methods cannot make use of any label information even though the class information of data is available. Although unsupervised methods can obtain the low-dimensional representation without using supervision, the performance may be degraded due to the unsupervised nature. As such, researchers have also investigated effective ways to extend CF to the fully-supervised/semi-supervised modes. One popular supervised CF variant called Supervised Graph Regularized Discriminative Concept Factorization (SGDCF) [22] uses the full class information of all input data to learn discriminative representations. Note that this kind of fully-supervised methods can only be used in limited application scenarios where all samples are labeled. However, class label information of samples is usually limited and the labeling process is also costly, so the study on fully-supervised CF methods is few. Instead, much more efforts have been paid to studying semi-supervised CF versions that can use less labeled data and a large number of unlabeled data for RL, which has broader application areas than supervised ones, such as semi-supervised image classification, text clustering, document retrieval and dimensionality reduction.

In this survey, we further divide the existing semi-supervised CF models into the following three sub-categories:

1) **Joint classification based CF.** The methods of this kind aim to learn a class indicator matrix and perform CF jointly, such as Discriminative CF (DCF) [23] and its variant called Hyper-graph regularized discriminative CF (HDCF) [24]. DCF and HDCF combine the data representation and data classification into a unified model, such that the discriminability ability can be strengthened significantly.

2) **Label constraint based CF.** This kind of methods generally design a label constraint matrix to represent label information and then incorporate it into CF. As a result, learned compact representation can be consistent with the known label information of data. Some classical methods of this type include Class-Driven CF (CDCF) [34], Constrained CF (CCF) [29] and its extensions like Local Regularization CCF (LRCCF) [30], Robust Semi-Supervised CF (RSSCF) [31], CCF with Graph Laplacian (CCF-GL) [32], Graph-based Discriminative CF (GDCF) [44] and Semi-supervised Discriminative CF (SDCF$_1$) [44]. Particularly, the recently proposed Robust Semi-Supervised Adaptive CF (RS$^2$ACF) algorithm [33] provides a new and more effective way to design the label constraint matrix by not only using the labeled data, but also predicting the label information of unlabeled samples.

3) **Pairwise constraint based CF.** Pairwise must-link and cannot-link constraints as used as additional constraints for semi-supervised CF, where the pairwise constraints are clearly defined based on the class information of labeled data. Classical semi-supervised CF methods consist of Pairwise Constrained CF (PCCF) [25] and its variants, including Semi-Supervised CF (SSCF) [26], Constrained Neighborhood Preserving CF (CNPCF) [27] and Regularized CF (RCF) [28].

It is worth noting that all aforementioned CF-based methods utilize the single-layer structure, i.e., the factorization process is performed only once. As a result, these single-layer CF methods can only discover "shallow" features, i.e., they cannot mine the deep information hidden in the data. With the fast development and great success of deep learning for the representation leaning and vision computing [165-169], researchers have also turned to study the deep/multi-layer CF methods to uncover the deep features and hierarchical structures embedded in data. Due to the huge challenge and difficulty of such research topic on the deep or multi-layer CF methods, there are only a few models that are proposed recently. To figure out the optimization strategies, we break the discussion down into two cases:

1) **Traditional feeding-style deep/multi-layer CF.** This kind of deep CF models is usually accumulating the layers simply. Because the most commonly-used approach of extending the single-layer CF models to deep ones is to feed the representation of previous layer as the input of the next layer directly. Several representative methods include Multilayer CF (MCF) [35], Graph Regularized MCF (GMCF) [36] and Dual-graph regularized MCF (DGMCF) [164]. However, such a strategy may be invalid in practice, since in fact the learned representation of the first layer decides the representation abilities of the whole framework by this way. However, existing models cannot ensure the output of the last layer to be a good representation, so directly feeding it to the next layer may degrade the performance of subsequent layers directly.

2) **Optimized deep/multi-layer CF.** Different from traditional feeding-style deep methods, these optimized methods clearly design novel hierarchical factorization architectures by using the multiple layers of linear transformations or updating the basis concepts/new representations in each layer to obtain the latent features through a progressive way. For this type of optimized deep CF methods, Deep Self-representative CF Network (DSCF-Net) [37] and Deep Semi-Supervised Coupled Factorization Network (DS$^2$CF-Net) [38] are two most frontier methods. Compared with DSCF-Net that updates the set of basis concepts to indirectly improve the representation result, DS$^2$CF-Net clearly designs a new deep coupled factorization architecture that can jointly update the basis concepts and new representation in each layer.

We outline the remainder of this survey paper as follows. Section II briefly reviews the root method of CF. We summarize the CF-based single-layer methods in Section III. In Section IV, we mainly describe the deep/multi-layer structures for CF. Section V conducts experiments to compare different CF methods. Section VI discusses the different applications of CF-based methods. Finally, the paper is concluded in Section VII, and we also provide some future directions for the research on CF.

## II. NOTATIONS AND REVIEW OF CF

### A. Important Notations

To facilitate the presentation and introduce the methods in this paper, we first list the important notations in Table I.

### B. Concept Factorization

CF, as a new variant of NMF, was proposed to characterize the possible nonlinear structure of samples. The regular CF method aims to represent each concept as a linear combination of data points and represent each data point as a linear combination of concepts. Given a data matrix $X = [x_1, x_2, ..., x_N] \in \mathbb{R}^{D \times N}$, where $x_i, i \in \{1, 2, ..., N\}$ is a sample vector. Let $U \in \mathbb{R}^{D \times r}$ and $V \in \mathbb{R}^{N \times r}$ be two nonnegative matrix factors whose product $UV^T \in \mathbb{R}^{D \times N}$ is the approximation to $X$. By representing each basis by a linear combination of $x_i$, i.e., $\sum_{i=1}^{N} w_{ij} x_i$, where $w_{ij} \geq 0$, CF proposes to solve the following minimization problem:

$$O_{CF} = \left\| X - XWV^T \right\|_F^2, \quad s.t. \ W, V \geq 0, \quad (1)$$

where $\left\| \cdot \right\|_F^2$ represents the squared Frobenius norm of a matrix,

Table I. Descriptions of Used Important Notations.

| Symbol | Description | Symbol | Description |
|---|---|---|---|
| $\mathbb{R}$ | Operating space | $r$ | The rank of matrix factorization |
| $X$ | The original data matrix $X=[x_1,x_2,...,x_N]\in\mathbb{R}^{D\times N}$ | $S$ | Graph weight matrix |
| $X_L$ | Labeled data matrix | $D_S$ | Diagonal matrix or degree matrix over $S$ |
| $X_U$ | Unlabeled data matrix | $L$ | Graph Laplacian matrix |
| $l$ | The number of labeled samples | $I$ | Identity matrix with compatible dimension |
| $u$ | The number of unlabeled samples | $\mathbf{1}$ | All-ones column vector |
| $x_i$ | The $i$-th sample vector of $X$ | $\mathbf{E}$ | All-ones matrix |
| $D$ | The original dimensionality of $X$ | $b$ | Bias vector |
| $N$ | The number of samples in $X$ | $M$ | Number of layers in deep network models |
| $U$ | One of CF factors: base matrix | $p$ | Number of nearest neighbors of each $x_i$ |
| $V$ | One of CF factors: new representation | $N_p(x_i)$ | The nearest neighbor set of each $x_i$ |
| $XW$ | Base matrix | $O$ | Objective function |
| $Z$ | Auxiliary matrix | $V^T$ | The transpose of matrix $V$ |
| $A$ | Label constraint matrix | $\sigma$ | Kernel width or kernel parameter |
| $E$ | Sparse error term | $\varepsilon$ | A small constant |

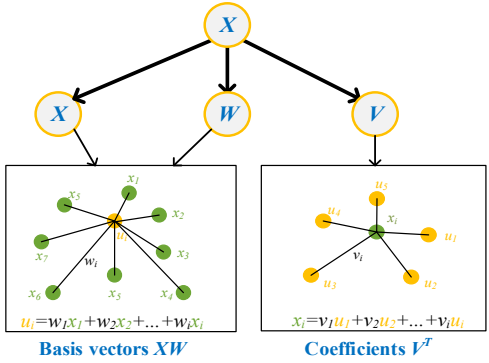

**Fig.2:** The framework of the standard CF algorithm.

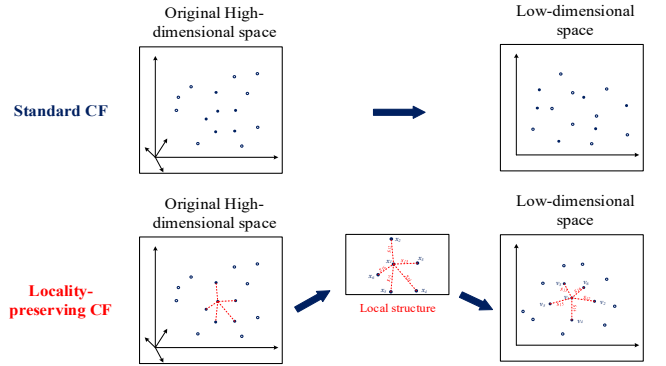

**Fig.3:** Comparison between the standard CF and locality-preserving CF methods.

$W=[w_{ij}]\in\mathbb{R}^{N\times r}$, $XW$ approximates the bases and $V^T$ denotes the new representation of $X$, which can be used for clustering. Note that CF solves Eq.(1) by the following updating rules:

$$w_{ik}^{t+1} \leftarrow w_{ik}^t \frac{(KV)_{ik}}{(KWV^TV)_{ik}},\quad v_{jk}^{t+1} \leftarrow v_{jk}^t \frac{(KW)_{jk}}{(VW^TKW)_{jk}}, \quad (2)$$

where $K=X^TX$ is inner product matrix. After the convergence, the new representation $V^T$ of original data can be obtained. The framework of CF is shown in Fig.2. We also summarize the optimization procedures of the root CF method in Algorithm 1.

---

**Algorithm 1: CF algorithm**
**Input:** Data matrix $X\in\mathbb{R}^{D\times N}$, rank $r$, a small constant $\varepsilon$.
**Initialization:** Construct the kernel matrix $K=X^TX$; Initialize $W$ and $V$ to be random matrices.
*While not converged do*
1. Update $w_{ik}^{t+1}$ by $w_{ik}^t(KV)_{ik}/(KWV^TV)_{ik}$;
2. Update $v_{jk}^{t+1}$ by $v_{jk}^t(KW)_{jk}/(VW^TKW)_{jk}$;
3. Convergence check: if $\|(O_{CF})^{t+1}-(O_{CF})^t\|_F \leq \varepsilon$, stop; else, return to step 1.
**Output:** New low-dimensional representation $V^*$ of $X$.

---

### III. SINGLE-LAYER CF-BASED METHODS

The CF-based single-layer methods are discussed in this section. According to how much prior class label information is used, we divide these shallow CF methods into three categories: i.e., unsupervised, supervised and semi-supervised ones.

#### A. Unsupervised CF Variants

Based on the basic CF method, lots of unsupervised methods have been proposed to improve the data representation and clustering powers. Specifically, researchers have made great efforts to improve the learnt new representations based on the following characteristics, i.e., locality, robustness, sparseness, operating space, discriminability and multiple-views. Accordingly, these CF methods are summarized as the locality-preserving CF methods, robust CF methods, kernel CF methods, discriminative CF methods and multi-view CF methods, respectively.

**1. Unsupervised Locality-Preserving CF Methods**

We know that the root CF method is a global model which can only preserve the global Euclidean geometry, however it cannot preserve the local manifold geometry [159]. To inherit the merits of CF and preserve the locality structure of data, many locality-preserving variants were proposed in recent years. Locality-preserving CF methods aim to keep the local structures in the original high-dimensional space during matrix decomposition, such that the learnt low-dimensional representation can have a more separable distribution than that of CF, which is benefit for subsequent clustering or recognition tasks. Fig.3 shows the difference between CF and locality-preserving variants in the working space. To retain the local information, we summarize the widely-used locality-preserving strategies in Fig.4. The first way is to incorporate the graph regularization into CF. Second, the local coordinate coding is also a popular way to extend CF to local scenario. The third one is to incorporate the self-representation learning into regular CF framework. Next, these locality-preserving strategies will be introduced.

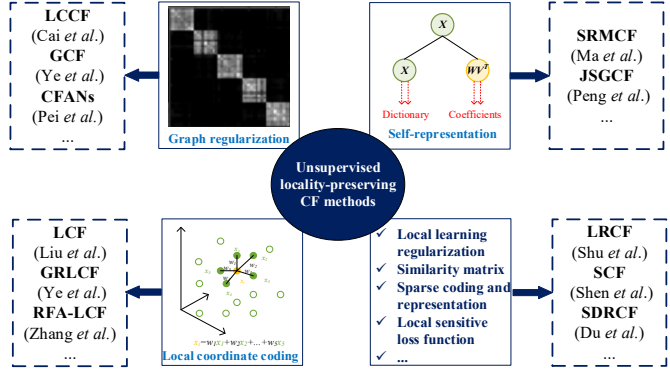

**Fig.4:** Different locality-preserving strategies and methods.

## 1.1 Graph-regularized CF methods

### 1.1.1 Artificial graph construction based CF variants:

We first review on the traditional graph regularization on CF. For the data matrix $X$, one can construct a graph with $N$ vertices where each vertex corresponds to a sample. Then one can obtain the weight matrix $S=[S_{ij}]$, $i,j \in \{1,2,...,N\}$ over the edges by the following five commonly-used ways:

**1) 0-1 weighting:**

$$S_{ij} = \begin{cases} 1 & \text{if } x_i \in N_p(x_j) \text{ or } x_j \in N_p(x_i) \\ 0 & \text{otherwise} \end{cases}, \quad (3)$$

where $N_p(x_i)$ denotes the $p$-nearest neighbor-set of sample $x_i$.

**2) Cosine similarity weighting:**

$$S_{ij} = \begin{cases} \dfrac{x_i^T x_j}{\|x_i\|\|x_j\|} & \text{if } x_i \in N_p(x_j) \text{ or } x_j \in N_p(x_i) \\ 0 & \text{otherwise} \end{cases}, \quad (4)$$

where $\|x_i\|$ denotes the $l_2$-norm of the sample vector $x_i$.

**3) Heat kernel weighting:**

$$S_{ij} = \begin{cases} \exp\left(-\dfrac{\|x_i-x_j\|^2}{2\sigma^2}\right) & \text{if } x_i \in N_p(x_j) \text{ or } x_j \in N_p(x_i) \\ 0 & \text{otherwise} \end{cases}, \quad (5)$$

where $\sigma$ denotes the kernel width or kernel parameter.

**4) Dot-product weighting:**

$$S_{ij} = \begin{cases} x_i^T x_j & \text{if } x_i \in N_p(x_j) \text{ or } x_j \in N_p(x_i) \\ 0 & \text{otherwise} \end{cases}. \quad (6)$$

---

**Algorithm 2: LCCF algorithm**
**Input:** Data matrix $X \in \mathbb{R}^{D \times N}$, rank $r$, a nonnegative parameter $\lambda_{LCCF}$ and a small constant $\varepsilon$.
**Initialization:** Construct kernel matrix $K = X^T X$; Initialize $W$ and $V$ to be random matrices; Construct the weight matrix $S$ and the diagonal matrix $(D_s)_{ii} = \sum_j S_{ij}$.
*While not converged do*
1. Update $w_{ik}^{t+1}$ by $w_{ik}^t (KV)_{ik} / (KWV^TV)_{ik}$;
2. Update $v_{jk}^{t+1}$ by
$v_{jk}^t (KW + \lambda_{LCCF} SV)_{jk} / (VW^TKW + \lambda_{LCCF} D_s V)_{jk}$;
3. Convergence check: if $\|(O_{LCCF})^{t+1} - (O_{LCCF})^t\|_F \leq \varepsilon$, stop; else, return to step 1.
**Output:** New low-dimensional representation $V^*$ of $X$.

Note that when the data point is normalized to 1, the dot-product weight of two normalized data points is identical to the cosine similarity weight.

**5) Reconstruction weight learning:** This weighting way is motivated by the Locally Linear Embedding (LLE) [138] that is a classical nonlinear dimensionality reduction method. LLE clearly provides a weighting scheme to preserve the neighborhood relationship. Specifically, LLE-style weighting method firstly assigns $p$ nearest neighbors for each sample $x_i$, and then computes the reconstruction weights $S_{ij}$ by solving the following constrained least-squares problem:

$$\min_{S_{ij}} \sum_i \left\| x_i - \sum_{x_j \in N_p(x_i)} S_{ij} x_j \right\|^2, \quad s.t. \sum_{x_j \in N_p(x_i)} S_{ij} = 1. \quad (7)$$

It is worth noting that the above artificial weighting methods have been proved to be effective to retain the local manifold structures of data, which have been widely-used in graph-regularized CF methods. In what follows, we will review the related graph-regularized CF methods clearly.

**LCCF [7].** Locally Consistent CF (LCCF) [7] was proposed in 2011, which is the first work to incorporate the graph regularization into CF. More specifically, LCCF defines the graph weight matrix $S$ using the cosine similarity weights over the $p$-nearest neighbor graph. After $S$ is obtained, one can obtain the graph Laplacian matrix $L = D_s - S$, where $D_s$ is a diagonal matrix with its entries being $(D_s)_{ii} = \sum_j S_{ij}$. The graph regularization can then be written as $tr(V^T L V)$. Finally, LCCF solves the following minimization problem:

$$O_{LCCF} = \|X - XWV^T\|_F^2 + \lambda_{LCCF} tr(V^T L V), \quad (8)$$
$$s.t. \ W, V \geq 0$$

where $\lambda_{LCCF}$ is an nonnegative regularization parameter. In this way, LCCF can extract the locality-preserving concepts and enhance the representation learning power over CF. The optimization procedure of LCCF is summarized in Algorithm 2. Note that LCCF is actually a general framework for graph-regularized CF, since we can use different artificial weighting schemes to compute the weight matrix, besides the cosine similarities.

**GCF [8].** Note that LCCF only considers one-sided clustering problem, i.e., it clusters data only depending on the similarities along features [7]. However, the co-clustering methods have been shown to be superior to traditional one-sided clustering. Thus, a new dual-graph regularized CF method termed GCF [8] was recently proposed for co-clustering. Specifically, GCF includes the graph regularizers of both the data manifold and feature manifold into CF simultaneously by constructing a $p$-nearest neighbor data graph $G^V$ and a $p$-nearest feature graph $G^U$. Then, GCF adopts the binary weighting scheme for constructing the dual-graphs $G^V$ and $G^U$, and then defines the weights $S^V$ and $S^U$ using binary weights as follows:

$$\begin{cases} (S^V)_{ij} = \begin{cases} 1 & \text{if } x_j \in N_p(x_i) \\ 0 & \text{otherwise} \end{cases}; i,j=1,2,...,N \\ (S^U)_{ij} = \begin{cases} 1 & \text{if } x_j \in N_p(x_i) \\ 0 & \text{otherwise} \end{cases}; i,j=1,2,...,D \end{cases}. \quad (9)$$

The graph Laplacian matrices over $G^V$ and $G^U$ can then be defined as $L^V = D^V - S^V$ and $L^U = D^U - S^U$, where $D^V$ and $D^U$ are diagonal matrices with entries being $(D^V)_{ii} = \sum_j (S^V)_{ij}$ and $(D^U)_{ii} = \sum_j (S^U)_{ij}$. The objective function of GCF is defined as

$$O_{GCF} = \|X - XWV^T\|_F^2 + \alpha_{GCF} tr(V^T L^V V) + \beta_{GCF} tr(W^T L^W W), \quad (10)$$

where $L^W = X^T L^U X$ denotes an auxiliary matrix, $\alpha_{GCF}$ and $\beta_{GCF}$ are two nonnegative trade-off parameters.

**ADGCF [13].** Most real data have noise, outliers and irrelevant features, so directly building the affinity graph based on the original input space as LCCF and GCF may be inaccurate due to the interference, which is in fact suffered by almost all the traditional graph regularized models. To alleviate this deficiency, the recently proposed ADGCF unifies the feature selection and dual-graph learning into the CF framework for joint optimization. Specifically, ADGCF performs feature selection and then uses the Gaussian kernel weighting scheme to construct the $p$-nearest neighbor data graph with the selected features in each iteration. By this way, it can successfully obtain optimized weights and can also ensure the optimized weights to be optimal for CF, but it still cannot avoid the tricky issue of determining the value of $p$ when choosing neighbors.

**MCFAW [11].** MCFAW employs a linear combination of multiple graphs to construct a graph regularizer and learns an optimal weight set for all graphs adaptively without introducing any additional parameter. The objective function of MCFAW is

$$O_{MCFAW} = \|X - XWV^T\|_F^2 + \alpha_{MCFAW} \sum_{i=1}^q \lambda_i^{AW} tr(V^T L_i^{AW} V), \quad (11)$$
$$s.t.\ W, V \geq 0,\ \lambda_i = 1/\left(2\sqrt{tr(V^T L_i^{AW} V)}\right)$$

where $q$ denotes the number of neighborhood graphs, $\lambda_i^{AW}$ is the weight of the $i$-th Laplacian graph. Clearly, MCFAW is a multi-graph regularized algorithm and can adaptively determine the weights for these $q$ graphs but for each graph, it still employs the traditional strategy as LCCF to build the weight matrix. As a result, it will suffer from the same problem as the traditional graph-based methods for the choice of $p$.

Besides the above mentioned methods, lots of graph-regularized CF methods have been recently proposed one after another. Wherein, GRLCF [9], LGCF$_1$ [10], RDCF [121], GCCF [124], Constrained Graph CF (CGCF) [129] and Sparse Constrained Manifold Regularized CF (SMCF) [135] that used the graph Laplacian in the same way as LCCF. Specifically, LGCF$_1$ and RDCF utilize the cosine similarity weighting to construct the weight matrix, while CGCF and SMCF define the weight matrix by '0-1' weighting. Structured discriminative CF (SDCF$_2$) [128] and Neighborhood Preserving CF (NPCF) [130] use the LLE-style weighting to preserve the neighborhood structures.

**Remarks.** It is noteworthy that all these artificial weighting strategies have an obvious drawback, i.e., there is a very tough issue to choose the optimal number of nearest neighbors (i.e., $p$) for different datasets. Moreover, the performance of graph regularized methods was verified to be sensitive to the choice of $p$ in experiments. Therefore, how to build the adjacency graph and define the weight matrix using a parameter-free way is undoubtedly a problem to be solved. In addition, these traditional graph-regularized methods usually learn the graph regularizer independently of the matrix factorization phase, however such operation cannot ensure the pre-obtained weights to be joint-optimal for subsequent representation learning [13-15] [125]. As such, the adaptive weighting strategy that does not to select the number of nearest neighbors has received some attention in the recent years. Specifically, several optimized adaptive graph regularized CF methods have been proposed.

### 1.1.2 Optimized graph construction based CF variants:

**CFANs [12].** CFANs was proposed in 2016, which used an adaptive neighbor weighting strategy to build the graph weight matrix $S^{ANs}$. Specifically, CFANs learns the neighbor connectivity of data by solving the following minimization problem:

$$\min_{S^{ANs}} \sum_{i=1}^N \left\{ \sum_{j=1}^N \left( \|x_i - x_j\|^2 S_{ij}^{ANs} + \frac{1}{2} \alpha_{ANs} \left(S_{ij}^{ANs}\right)^2 \right) \right\}, \quad (12)$$
$$s.t.\ S^{ANs} \geq 0,\ \left(S_{i.}^{ANs}\right)^T \mathbf{1} = 1,\ S_{ii}^{ANs} = 0$$

where $\alpha_{ANs}$ is a regularization parameter and $\mathbf{1}$ is a column vector with all ones. Note that if $\alpha_{ANs} = 0$, it will lead to a trivial solution of Eq.(12), i.e., only the nearest data point can be a neighbor of $x_i$ with weight 1. If $\alpha_{ANs}$ is infinite, then the optimal solution of Eq.(12) is that all of the data points are neighbors of $x_i$ with the weight $1/N$. That is, the neighbors can be assigned adaptively by tuning $\alpha_{ANs}$ according to [42]. After the weights are obtained, CFANs can obtain the graph Laplacian matrix by $L^{ANs} = D^{ANs} - \left(S^{ANs} + \left(S^{ANs}\right)^T\right)/2$, where $(D^{ANs})_{ii} = \sum_j S_{ij}^{ANs}$. Finally, by combining the neighbor graph regularization constraint and adaptive neighbor weights learning into CF as a united framework, CFANs solves the following problem:

$$O_{CFANs} = \|X - XWV^T\|_F^2 + \sum_{i=1}^N \left\{ \sum_{j=1}^N \left( \|x_i - x_j\|^2 S_{ij}^{ANs} + \frac{1}{2} \alpha_{ANs} \left(S_{ij}^{ANs}\right)^2 \right) \right\}$$
$$+ \beta_{ANs} tr(V^T L^{ANs} V) + \gamma_{ANs} \|V^T V - I\|,$$
$$s.t.\ W, V, S^{ANs} \geq 0,\ \left(S_{i.}^{ANs}\right)^T \mathbf{1} = 1,\ S_{ii}^{ANs} = 0$$

(13)

where $\beta_{ANs}$, $\gamma_{ANs}$ are nonnegative parameters and the penalty term $\|V^T V - I\|$ is to relax the orthogonal constraint $V^T V = I$.

It is worth noting that although CFANs can learn the weight matrix by adaptively assigning neighbors, it is still not parameter-free, since it needs to tune the number of neighbors when updating the regularization parameter $\alpha_{ANs}$ [12][42].

**CF-OGL [126] and GRLCF$_{CLR}$ [9].** To optimize the graph learning for CF, CF-OGL [126] and GRLCF$_{CLR}$ [9] were recently proposed. Where the Constrained Laplacian Rank (CLR) [43] is a new graph learning method proposed in 2016. Specifically, GRLCF$_{CLR}$ employs CLR to build the weight matrix, since CLR can learn a graph with exactly $\vartheta$ connected components, where $\vartheta$ is the number of clusters. Note that the weight matrix $S^{CLR}$ is obtained by solving the following problem:

$$\min_{S^{CLR}} \left\| S^{CLR} - S_0^{CLR} \right\|^2, \\ s.t. \sum_j \left( S^{CLR} \right)_{ij} = 1, \left( S^{CLR} \right)_{ij} \geq 0, rank\left( L^{CLR} \right) = N - \vartheta \quad (14)$$

where $S_0^{CLR}$ is an initial matrix, $L^{CLR} = D^{CLR} - \left( \left( S^{CLR} \right)^T + S^{CLR} \right) / 2$, $D^{CLR}$ is a diagonal matrix with its diagonal entries defined as $\left( D^{CLR} \right)_{ii} = \sum_j \left( S^{CLR} \right)_{ij}$. Note that although the CLR based methods can avoid choosing the value of $p$, it needs to determine the value of $\vartheta$, which also involves a tough selecting issue.

**RLLDCF [125].** To obtain a robust adaptive weight matrix for improving the representation ability, RLLDCF was recently proposed in 2018. Specifically, by denoting the all-ones matrix by $\mathbf{E}$, RLLDCF seeks the weight matrix $S^{RLL}$ by minimizing the following problem for robust local learning:

$$\left\| V^T - V^T S^{RLL} \right\|_{2,1} + \mu_{RLL} tr\left( \mathbf{E} S^{RLL} \right), \quad (15)$$

where $\mu_{RLL}$ is a regularization parameter and the $L_{2,1}$-norm $\left\| \cdot \right\|_{2,1}$ can ensure the reconstruction error to be robust to outliers and noise. Then, RLLDCF repeats learning the representation $V^T$ and the sparse weight matrix $S^{RLL}$ iteratively to capture the true geometrical structure of the data distribution adaptively. Besides, minimizing $tr\left( \mathbf{E} S^{RLL} \right)$ can enhance the sparsity of the solution. However, the above formulation may suffer from a trivial solution $S^{RLL} = I$, resulting in meaningless solution.

**RFA-LCF [14-15].** In order to realize the real parameter-free adaptive weight learning and avoid the trivial solution on the weight matrix, a novel auto-weighting mechanism was introduced in RFA-LCF [14-15]. Note that this weighting strategy differs from the other existing ones that: 1) it uses a $L_{2,1}$-norm based sparse projection $P \in \mathbb{R}^{D \times D}$ to map the original data into noise-removed clean data $P^T X$ and then factorizes data based on $P^T X$; 2) to encode the neighborhood information and pairwise similarities accurately, it retains the manifold structures jointly over $P^T X$, basis vectors $XW$ and new coordinates $V^T$ in an adaptive manner by minimizing the joint reconstruction error $\left\| P^T X - P^T X S^{SA} \right\|_F^2 + \left\| W - W S^{SA} \right\|_F^2 + \left\| V^T - V^T S^{SA} \right\|_F^2$, where $S^{SA}$ denotes a shared adaptive reconstruction weight matrix. Different from RLLDCF, RFA-LCF avoids the trivial solution $S^{SA} = I$ by forcing $\left( S^{SA} \right)_{ii} = 0$. In this way, RFA-LCF can update $S^{SA}$ adaptively in each iteration without specifying additional parameter. By incorporating the joint reconstruction error into CF, the objective function of RFA-LCF can be defined as follows:

$$O_{RFA-LCF} = \left\| X^T P + \mathbf{1} b^T - V W^T X^T \right\|_{2,1} + \alpha_{RFA-LCF} f(W, V) \\ + \beta_{RFA-LCF} \left( \left\| P^T X - P^T X S^{SA} \right\|_F^2 + \left\| W - W S^{SA} \right\|_F^2 + \left\| V^T - V^T S^{SA} \right\|_F^2 \right), \quad (16) \\ + \gamma_{RFA-LCF} \left\| P \right\|_{2,1}, \quad s.t. \quad W, V, S^{SA} \geq 0, \left( S^{SA} \right)_{ii} = 0, i = 1, 2, ..., N$$

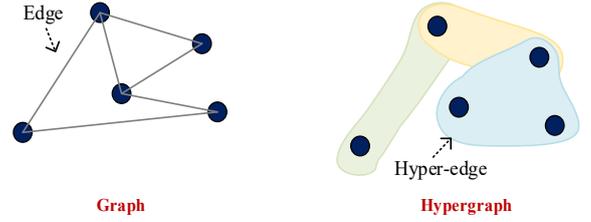

**Fig.5:** Difference between traditional graph and hyper-graph.

where $\alpha_{RFA-LCF}$, $\beta_{RFA-LCF}$ and $\gamma_{RFA-LCF}$ are nonnegative parameters. $\mathbf{1} b^T$ is added to relax the reconstruction error for avoid the overfitting. v is the robust adaptive locality and sparsity constraint term, which will be introduced later in this paper.

### 1.1.3 Hyper-graph construction based CF variants:

In real applications, the information among samples is critical, so the traditional weighting may be inaccurate. To overcome this issue and extract multi-geometry information, recent works add the weighted hyper-graph Laplacian regularization into CF.

The difference between the traditional graph and hyper-graph is shown in Fig.5. Specifically, in the traditional graph, an edge connects only two vertices, but in hyper-graph a hyper-edge can connect any number of vertices. Thus, hyper-graph can be used to describe more complex relationships than traditional graph. Representative hyper-graph methods consist of HDCF [24], Hyper-graph regularized CF (HRCF) [133], Hyper-graph Dual Regularization CF (DHCF) [134], and Local and global regularized CF (LGCF$_2$) [139]. In general, these hyper-graph based CF methods define the graph Laplacian as $L^{hyper} = D^{hyper} - S^{hyper}$, where $D^{hyper}$ is a diagonal matrix with entries being the vertex degree. The weight matrix $S^{hyper}$ is defined as $S^{hyper} = \hat{F} \hat{B} D_e^{-1} \hat{F}^T$, where $\hat{F}$ is the incidence matrix of the hyper-graph, $\hat{B}$ is a weight matrix with $\hat{B}_{ij}$ being the weights between any two vertices in hyper-graph and $D_e$ denotes a diagonal matrix whose entries denote the degrees of hyper-edges.

## 1.2 Local coordinate coding based CF methods

It is worth noting that the graph-regularized CF methods can obtain locality-preserving representations successfully, but may not always satisfy the sparsity conditions in certain practical application scenarios. To obtain both the locality and sparsity simultaneously, one category of CF methods based on local coordinate coding (LCC) has been developed.

**LCF [16-17].** LCF firstly incorporates the idea of local coordinate coding into CF. Specifically, LCF takes the locality constraint into account, considers the anchor points $u_j = \sum_i w_{ij} x_i$ and the coordinates for the sample $x_i$ over each column of $V^T$ with respect to the anchor points. Then, LCF defines the following constraint term to measure the locality and sparsity penalties between the anchor point $u_r$ and the sample $x_i$:

$$\sum_{\tau=1}^r \left| v_{\tau i} \right| \left\| u_\tau - x_i \right\|^2 = \sum_{\tau=1}^r \left| v_{\tau i} \right| \left\| \sum_{j=1}^N w_{j\tau} x_j - x_i \right\|^2. \quad (17)$$

Finally, LCF minimizes the following objective function:

$$O_{LCF} = \left\| X - XWV \right\|_F^2 + \lambda_{LCF} \sum_{i=1}^N \sum_{\tau=1}^r \left| v_{\tau i} \right| \left\| \sum_{j=1}^N w_{j\tau} x_j - x_i \right\|^2, \quad (18)$$

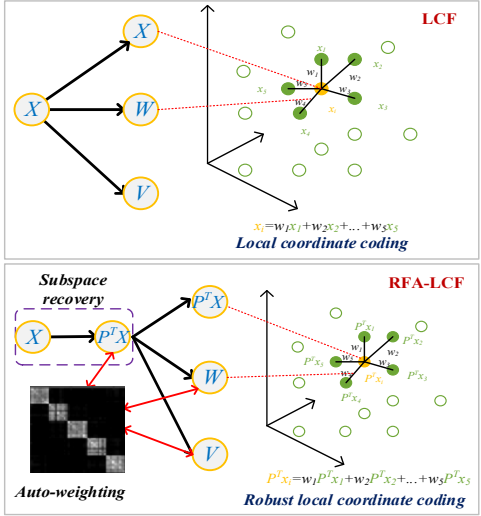

**Fig.6:** Difference between LCF and RFA-LCF.

where $\lambda_{LCF} \geq 0$ is a weighting parameter. Based on Eq.(17), LCF can represent each sample $x_i$ by using only a few nearby anchor points so that the sparsity and local structure can be retained.

However, LCF fails to preserve the manifold structure in the data space and its locality constraints fail to reveal the intrinsic data structure as well. To address this issue, recent methods also are developed to jointly consider the local geometric structures of the data manifold and the local coordinate coding as additional constraints. Several representative methods include GLCF [18], LGCF$_1$ [10], GRLCF [9], GRLCF$_{CLR}$ [9], and RFA-LCF [14-15], etc. Specifically, GLCF adds the manifold kernel learning into LCF model to reveal the semantic structure in the warped RKHS. LGCF$_1$, GRLCF and GRLCF$_{CLR}$ are all based on the combinations of the graph-regularization and local coordinate coding, which aims at addressing the following minimization based optimization problem:

$$\|X - XWV^T\|_F^2 + \chi \sum_{i=1}^{N}\sum_{\tau=1}^{r}|v_{\tau i}^T|\|\sum_{j=1}^{N}w_{j\tau}x_j - x_i\|^2 + \delta tr(V^TLV),$$
$$s.t.\ W, V \geq 0$$
(19)

where $\chi$ and $\delta$ denote two tunable trade-off parameters.

It is noted that the only difference between LGCF$_1$, GRLCF and GRLCF$_{CLR}$ is that they construct the graph weights by using different approaches. Specifically, LGCF$_1$ and GRLCF define the weight matrix similarly as LCCF based on the $p$-nearest-neighbor adjacency graph, while GRLCF$_{CLR}$ clearly encodes the graph weights by using the CLR algorithm [43].

RFA-LCF improves LCF in three aspects: 1) RFA-LCF performs the CF on the recovered clean data $P^TX$ as mentioned above; 2) RFA-LCF includes the auto-weighting scheme into CF to better preserve the local geometrical structures by minimizing the joint neighborhood preserving error; 3) RFA-LCF optimizes the local coordinate coding term by a robust locality and sparsity constraint term $f(W,V)$, and uses it in the objective function in Eq.(9), which is defined as follows:

$$f(W,V) = \sum_{i=1}^{N}\sum_{r=1}^{R}|v_{ri}^T|\|\sum_{j=1}^{N}w_{jr}P^Tx_j - P^Tx_i\|^2.$$
(20)

That is, the local coordinate coding process of RFA-LCF is operated in a recovered clean feature space rather than in the original data space that usually contains various noise. Note that this is clearly different from LCF that directly encodes the local coordinates in the original data space. The difference between LCF and RFA-LCF has been illustrated in Fig.6.

### 1.3 Self-representation based CF methods

Different from the graph regularized and LCC based CF methods, the self-representation based CF methods constructs the affinity matrix using the new representation rather than the original raw data $X$. Specifically, they consider $WV^T$ as a coefficient matrix based on the dictionary of data $X$, which is also regarded as a meaningful representation of $X$. Representative methods of this kind include SRMCF [20] and JSGCF [21].

**SRMCF [20].** SRMCF incorporates the self-representation with the adaptive neighbor structure as [12][42] to assign the neighbors for all samples. By integrating the adaptive neighbor structure and manifold regularizer into CF, the objective function of SRMCF is defined as follows:

$$O_{SRMCF} = \|X - XWV^T\|_F^2 + \beta_{SRMCF} tr(V^TLV)$$
$$+ \alpha_{SRMCF}\sum_{i=1}^{N}\left\{\sum_{j=1}^{N}\left(\|(WV^T)_i - (WV^T)_j\|^2 S_{ij}^{ANs} + \frac{1}{2}\alpha_{ANs}(S_{ij}^{ANs})^2\right)\right\},$$ (21)
$$s.t.\ W,V,S^{ANs} \geq 0,\ (S_{i.}^{ANs})^T \mathbf{1} = 1,\ 1 \geq S_{i.}^{ANs} \geq 0$$

where $S_{ij}^{ANs}$ denotes the probability of $x_j$ (excluding itself from $X$) being connected to $x_i$ as a neighbor, $\alpha_{SRMCF}$ and $\beta_{SRMCF}$ denote the regularization parameters. The constraints $(S_{i.}^{ANs})^T \mathbf{1} = 1$ and $1 \geq S_{i.}^{ANs} \geq 0$ can ensure the probability property of $S_{i.}^{ANs}$. $L$ is a predefined graph Laplacian matrix by using the '0-1' weighting based on the Euclidean distances of samples LCCF [7].

**JSGCF [21].** JSGCF is motivated by the fact that jointly performing the structured graph learning and clustering can avoid the suboptimal solutions caused by the two-stage strategy in the traditional graph learning. Suppose that the data points have $\vartheta$ clusters, JSGCF aims to build a graph to guide the data points to be divided into $\vartheta$ clusters without any post-processing. Thus it imposes the rank constraint on the graph Laplacian matrix $L_{WV^T}$ of $WV^T$ as $rank(L_{WV^T}) = N - \vartheta$. The rank constraint can be converted to an equivalent mathematical expression:

$$\min_{F \in \mathbb{R}^{N \times \vartheta}, F^TF = I} tr(F^TL_{WV^T}F),$$ (22)

where each row $f_i$ of $F$ can be seen as a vector connected to data point $x_i$ on the graph $WV^T$. For the convenience of optimization, JSGCF introduces an auxiliary matrix $\Theta$ to approximate $WV^T$ and finally solves the following problem:

$$O_{JSGCF} = \|X - XWV^T\|_F^2 + \alpha_{JSGCF} tr(F^TL_\Theta F) + \beta_{JSGCF}\|\Theta - WV^T\|_F^2,$$
$$s.t.\ W,V,\Theta \geq 0,\ \Theta\mathbf{1} = 1,\ F \in \mathbb{R}^{N \times \vartheta},\ F^TF = I$$
(23)

where $\alpha_{JSGCF}$ and $\beta_{JSGCF}$ are two tunable trade-off parameters.

### 1.4 Other Locality-preserving CF methods

In additional to the above-mentioned locality-preserving methods, there are also some other techniques to preserve the locality of samples during the concept factorization process.

**LRCF [30].** LRCF preserves the local information by incorporating the local learning regularization. That is, LRCF partitions the input data space into some local regions and minimizes the predicting cost over each region. Suppose that input data has $c$ classes, LRCF defines a predicting function $f_i^l(x)$ in $N(x_i)$ to estimate the cluster label of $\{x_j\}_{x_j \in N(x_i)}$, where $1 \le l \le c$ and $N(x_i)$ is the neighborhood of $x_i$. Then, the minimization of the overall local prediction costs can be defined as follows:

$$\min \sum_{l=1}^{D} \sum_{i=1}^{N} \|f_i^l(x_i) - v_i^l\|^2, \quad (24)$$

Then, the local learning regularization can be defined in matrix trace form as $tr(V^T M_{LRCF} V)$, where $M_{LRCF} = (S' - I)^T (S' - I)$. $S'_{ij}$ is equal to $\delta_{ij}$ if $x_j \in N(x_i)$, and otherwise 0. Finally, LRCF solves the following minimization problem:

$$O_{LRCF} = \|X - XWV^T\|_F^2 + \alpha_{LRCF} tr(V^T M_{LRCF} V), \text{ s.t. } W, V \ge 0, \quad (25)$$

where $\alpha_{LRCF}$ denotes a nonnegative trade-off parameter.

**SCF [127] and RSCF [127].** To enforce the reconstructed samples by CF to be close to that of original samples, two similarity-based models, i.e., SCF and RSCF, are proposed, which include a similarity matrix into CF. Denote the similarity matrix by $K^S$ of $N$ samples, this similarity reconstruction can be represented as $K^S \approx V^T V$. SCF constructs $K^S$ using classical $p$-nearest neighbor to encode the similarity as [140]. To combine the data reconstruction and similarity reconstruction into a unified reconstruction error term, SCF finally solves the following coupled optimization problem:

$$O_{SCF} = \|K^S - VW^T K^S WV^T\|, \text{ s.t. } W, V \ge 0. \quad (26)$$

Note that RSCF is just the robust extension of SCF, which will be depicted later in this paper.

**SDRCF [120].** SDRCF is inspired by the sparse representation (SR) [56-66], which simultaneously incorporates the local geometrical structures of both the data and features into CF, and obtain a weight matrix. For a sample $x$ and a matrix $\mathcal{D} \in \mathbb{R}^{D \times N}$ containing the dictionary atoms in its columns, SR represents $x$ using as few entries of $\mathcal{D}$ as possible, defined as follows:

$$\min_{s^{SR}} \|s^{SR}\|_1, \text{ s.t. } x = \mathcal{D} s^{SR}, \quad (27)$$

where $s^{SR}$ is the sparse coefficient and $\|\cdot\|_1$ is the $L_1$-norm of a vector. In this way, SDRCF can obtain a sparse weight matrix $S^{SR}$. Then, the graph Laplacian matrices $L^V$ and $L^W$ can be similarly obtained as GCF [8] based on $S^{SR}$. Finally, the objective function of SDRCF can be formulated as follows:

$$O_{SDRCF} = \|X - XWV^T\|_F^2 + \alpha_{SDRCF} tr(V^T L^V V) + \beta_{SDRCF} tr(W^T L^W W), \quad (28)$$
$$\text{s.t. } W, V \ge 0$$

where $\alpha_{SDRCF}$ and $\beta_{SDRCF}$ are parameters trading-off the graph-regularization on the data and features, respectively. Note that though SDRCF learns the graph weight matrices by SR, it is essentially a graph-based CF model.

**LSDCL [183].** Due to the fact that Correntropy Induced Metric (CIM) is more suitable to characterize the locality of samples than the traditional $L_2$-norm used in LLE-style weighting, LSDCL introduces CIM as a generalized metric based on information-theoretic learning (ITL), which is defined as

$$\mathcal{L}(e) = 1 - \exp\left(-\frac{e^2}{\sigma^2}\right), \quad (29)$$

where $\sigma$ is the kernel width as in the Gaussian function, and $e$ denotes reconstruction error. Then LSDCL adopts the local sensitive loss function, which can be formulated as follows:

$$\min_V \sum_{i=1}^{N} \sum_{j=1}^{N} S_{ij} \left(1 - \exp\left(-\|V^T x_i - V^T x_j\|^2 / \sigma^2\right)\right), \quad (30)$$

where $S_{ij}$ is the 0-1 weight for $x_i$ and $x_j$. The local sensitive loss function emphasizes more on most similar pairs with small errors to better characterize the local structure of data [183].

**CFLCs [186].** From the perspective of clustering data, the columns of basis vectors $U=XW$ can be seen as clustering centroids in CF-based methods. Based on the fact that in real world, each class of data may contain several sub-classes, CFLCs supposes that there are $r$ ($c < r < N$) local centroids. First, CFLCs considers $w_{ij}$ as the probability that $x_i$ is associated with the local centroid $u_j$, thus it can preserve the locality information by minimizing $\sum_{i=1}^{N} \sum_{j=1}^{r} w_{ij} \|x_i - u_j\|_2^2$ with the constraint $\sum_i w_{ij} = 1$. Second, to obtain clear clustering structure, CFLCs denotes a bipartite graph as $\tilde{W} = [\mathbf{0} \ W; W^T \ \mathbf{0}]$. To make the bipartite graph have exact $c$ components, the rank of the Laplacian matrix $L_W$ must be $N+r-c$. To enforce the rank of $L_W$ to be $N+r-c$, CFLCs includes a minimization term $\min tr(\mathcal{P}^T L_W \mathcal{P})$ into the standard CF framework, where $\mathcal{P} \in \mathbb{R}^{(N+r) \times c}$ is the auxiliary matrix and $\mathcal{P}^T \mathcal{P} = I$. In this way, the optimal $W$ is able to connect the samples with appropriate local centroids and the bipartite graph $\tilde{W}$ can indicate the clustering results directly [186].

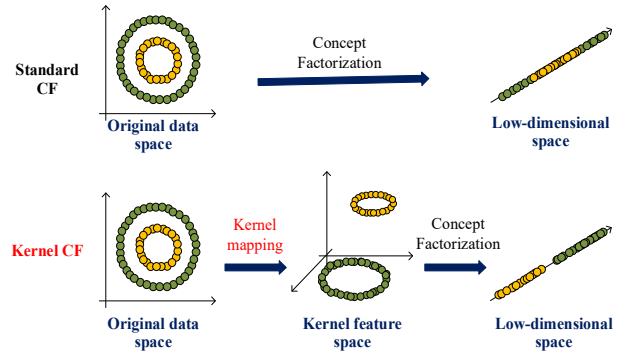

**Fig.7:** Visual comparison of performing CF in the original Euclidean space and kernel feature space.

## 2. Unspervised Kernel CF Methods

Kernel CF is mainly motivated by the fact that original linearly-inseparable samples may be linearly-separable if after being mapping a higher-dimensional RKHS using the kernel trick. As a result, the performance of data reconstruction, clustering or recognition tasks can be potentially improved for the linear separability. Similarly, kernel CF assumes that the factorization process in kernel feature space will also be more accurate. Note that Fig.7 is the visual comparison of performing CF in original Euclidean space and kernel feature space.

**MKCF [19].** The main advantages of CF over NMF is that it can be kernelized for discovering the nonlinear structures hidden in data. By considering the manifold kernel learning into CF, MKCF was recently proposed. First, the objective function of CF can be rewritten in the form of matrix trace as

$$O_{CF} = tr\left((I-WV^T)^T X^T X (I-WV^T)\right), \quad (31)$$

where $X^T X$ is the inner product of matrices. Then, a generalized formulation of kernel CF (KCF) can be expressed as

$$\begin{aligned}O_{KCF} &= \|\phi(X) - \phi(X)WV^T\|_F^2 \\ &= tr\left((I-WV^T)^T \langle\phi(X),\phi(X)\rangle (I-WV^T)\right),\end{aligned} \quad (32)$$

where the inner product $\langle\phi(X),\phi(X)\rangle = \phi(X)^T \phi(X)$ can be denoted by a kernel $\mathcal{K}$. $\phi(X)$ is a function which maps $X$ into a high-dimensional kernel space. By involving manifold learning into Eq.(32), MKCF finally solves the following problem:

$$\begin{aligned}O_{MKCF} &= tr\left((I-WV^T)^T \tilde{\mathcal{K}}_{MK} (I-WV^T)\right) \\ &= tr\left(\tilde{\mathcal{K}}_{MK} - 2VW^T\tilde{\mathcal{K}}_{MK} + VW^T\tilde{\mathcal{K}}_{MK}WV^T\right),\end{aligned} \quad (33)$$

where $\tilde{\mathcal{K}}_{MK} = \mathcal{K} - \mathcal{K}^T(I+L_{MK}\mathcal{K})^{-1} L_{MK}\mathcal{K}$ and $L_{MK}$ is the graph Laplacian matrix defined similarly as that of LCCF. Note that MKCF constructs the adjacency graph by $p$-nearest neighbor search and defines the weights of the neighborhood graph by using the binary weights. That is, MKCF also suffers from the tricky issue of selecting the optimal number of $p$.

**GLCF [18].** To further enhance the locality and sparsity of the learnt representations, GLCF also incorporates the local coordinate coding into kernel CF model. Thus, the objective function of GLCF can be formulated as follows:

$$\begin{aligned}O_{GLCF} = &tr\left((I-WV^T)^T \tilde{\mathcal{K}}_{M(:,:)} (I-WV^T)\right) \\ &+ \alpha_{GLCF} \sum_{i=1}^{N} tr\left(\mathbf{1}^T \Lambda_i \mathbf{1} \tilde{\mathcal{K}}_{MK(i,i)} - 2\cdot\mathbf{1}^T \Lambda_i W^T \tilde{\mathcal{K}}_{MK(:,i)} + W\Lambda_i W^T \tilde{\mathcal{K}}_{MK(:,:)}\right),\end{aligned} \quad (34)$$

where the parameter $\alpha_{GLCF} \geq 0$, $\Lambda_i = diag(|v_{i1}|,|v_{i2}|,...,|v_{ir}|) \in \mathbb{R}^{r\times r}$.

**GMKCF [136].** Although both MKCF and GLCF involves the kernel trick for nonlinear representation learning, they are single-kernel methods. Note that in the unsupervised scenario, it is a tough problem to select proper kernel function for a specified dataset in practical applications. To solve this problem, a Globalized Multiple Kernel CF (GMKCF) [136] was proposed in 2019. GMKCF mainly aims to provide multiple candidate kernel functions at the same time and performs concept factorization learning based on global linear fusion. Specifically, suppose there are $n_\eta$ kernels, then GMKCF defines the global kernel as $\mathcal{K}_\eta = \sum_{i=1}^{n_\eta} \mathcal{K}^i$, s.t. $\sum_{i=1}^{n_\eta} \theta_i = 1$, $\theta_i \geq 0$. Then, GMKCF addresses the following objective function:

$$O_{GMKCF} = \|\mathcal{K}_\theta - \mathcal{K}_\theta WV^T\|_F^2, \ s.t.\ W, V^T \geq 0, \sum_{i=1}^{n_\eta} \theta_i = 1, \theta_i \geq 0. \quad (35)$$

**DMKCF [182].** To alleviate the problem of kernel selection, DMKCF also employs the multiple-kernels strategy. Similar to GMKCF [136], DMKCF also assumes that there are $n_\eta$ kernel functions, and for each kernel function $\mathcal{K}^\eta$, it assigns a weight $\mu_\eta$ for $\mathcal{K}_\eta$. Finally, the combined kernel function can be defined as $\tilde{\mathcal{C}}(x_i,x_j) = \sum_{i=1}^{n_\eta} \mu_i^2 \mathcal{K}^i(x_i,x_j)$. Then, DMKCF solves the multiple kernel concept factorization problem by replacing the single kernel in Eq. (32) with the combined kernel.

## 3. Unsupervised Robust CF Methods

Noise and outliers are common in most real data in emerging apllications, as shown in Fig.8, which may degrade subsequent tasks. Therefore, performing CF directly on original raw data may not be a good choice, since the included noise may lead to abnormal data distribution resutling in unreliable representation results. Toward this issue, researchers have explored effective strategies to enhance the robustness of the CF variants to noise, i.e., recovering the underlying subspace prior to performing the factorization process. Representative methods include:

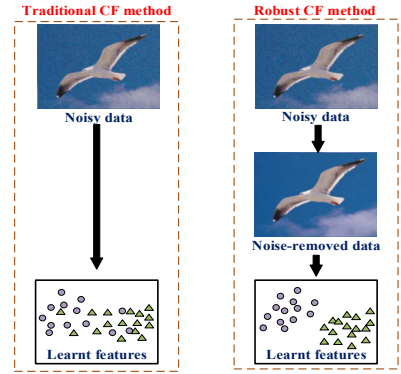

**Fig.8:** Traditional CF model *vs.* robust CF model.

**ADGCF [13].** To handle noisy and irrelevant data, ADGCF adopts a the feature selection (FS) to assign different weights for various features, so that the samples can be represented in a more proper way than directly applying the original features. Specifically, CF with FS (CF$_{FS}$) can be formulated as

$$O_{CF_{FS}} = \|diag(\xi)(X - XWV^T)\|_F^2, \quad (36)$$

where $diag(\xi)$ is a $D \times D$ diagonal matrix with the entries being the feature weight vector $\xi$. Then, ADGCF constructs two graphs adaptive to the selected features based on the new feature space defined by the feature weight vector $\xi$.

**RFA-LCF [14] and RLLDCF [125].** Tradtional CF-based methods usually make use of the Frobenius-norm to encode the reconstruction error of the matrix factorization. However, the Frobenius-norm is very sensitive to noisy data. Note that $L_{2,1}$-norm has been proven to be robust to noise and outliers [153-155], so it can be used to replace the Frobenius-norm to improve the robustness. Specifically, RLLDCF defines the robust CF model as $\min \|X - XWV^T\|_{2,1} + \|V^T\|_{2,1}$, where the first term is the $L_{2,1}$-norm constrained reconstruction error and the second term can make $V^T$ sparse in rows. RFA-LCF [14] introduces a $L_{2,1}$-norm regularized projection $P$ to recover the underlying subspaces and remove noise from data. Specifically, the sub-problem of the robust subspace recovery in RFA-LCF can be defined as $\min \|X^T P - VW^T X^T\|_{2,1} + \|P\|_{2,1}$. Besides, RFA-LCF also learns the adaptive reconstruction weights based on the removed clean data $X^T P$ rather than original input data, so that the robustness and locality of the model can be further enhanced.

**RDCF [121].** Movited by the robust PCA (RPCA) [142] that decomposes a data matrix to a low-rank part and a sparse part, RDCF also assumes that the reconstruction $XWV^T$ is low-rank and the sparse part contains noise. Denote the sparse error by $\Gamma$, the robust sub-problem of RDCF is formulated as follows:

$$\min_{W,V,\Gamma} \|X - \Gamma - XWV^T\|_F^2 + \|\Gamma\|_1, \quad (37)$$

where $\|\Gamma\|_1 = \sum_{ij}|\Gamma_{ij}|$ is to guarantee the sparseness of $\Gamma$. In this way, the sparse error can be separated from the raw data.

**RSCF [127].** RSCF is a robust version of SCF. Different from SCF using the Frobenius-norm to measure the difference between the similarities of original data and reconstructed data, RSCF employs the Chebyshev norm $L_\infty$ [143] to replace the Frobenius-norm. Formally, RSCF proposes to minimize $\|\tilde{\mathcal{Z}}\|_\infty$, where $\mathcal{Z} = K^S - VW^T K^S WV^T$ in Eq.(26), and $\tilde{\mathcal{Z}}$ is the vectorized representation of the matrix $\mathcal{Z}$.

**GCCF [124].** To imrpove the robustness of resulted model to noise, GCCF provides another way to replace the tradtional Euclidean norm. Since correntropy has been proved to be able to deal with the non-Gaussian noise and outliers, GCCF uses correntropy as the similarity measure to calculate the distance between the original data and its recosntruction. Specifically, based on the maximum corrent criterion (MCC), GCCF solves the following maximization problem:

$$\max_{W,V} O_{GCCF} = \sum_{i=1}^{D} \varsigma\left(\sqrt{\sum_{j=1}^{N}\left(X_{ij} - (XWV^T)_{ij}\right)^2}\right) - \alpha_{GCCF} tr(V^T LV), \quad (38)$$

s.t. $W, V \geq 0$

where $\alpha_{GCCF} \geq 0$ is a tunable parameter, $\varsigma(a-b) \triangleq \exp^{-(a-b)^2/2\sigma^2}$.

## 4. Unspervised Discriminative CF Methods

Most unsupervised CF methods mainly focus on retaining the neighborhood structures of samples, but usually neglects the discriminative information. But discriminative information is very important for improving the clustering and representation learning performance of CF methods, especially in the absence of label information. The related models are discussed below:

**RDCF [121].** To improve the discriminative ability of the learned representation $V$, RDCF first introduces a $N \times r$ group indicator matrix $\Phi$, where $\Phi_{ij} = 1$ if the $i$-th sample belongs to the $j$-th group and otherwise $\Phi_{ij} = 0$. The scaled indicator matrix with respect to $\Phi$ is defined as $\mathcal{F} = \Phi(\Phi^T\Phi)^{-1/2}$. Although we did not know $\mathcal{F}$ in prior as we have no label information, we can still find that $\mathcal{F}^T\mathcal{F} = I$. To make the representation $V$ to characterize the discriminative structure in $\mathcal{F}$, RDCF forces them to be close to each other, i.e., minimizing the difference between them. Since $\mathcal{F}$ is orthogonal, RDCF enforces $V$ to be approximately orthogonal as $\|V^TV - I\| \leq \varepsilon$, where $\varepsilon$ is a small value. By using this approximate orthogonal constraint, RDCF can effectively capture the discriminative information in data.

**RLLDCF [125].** To discover the discriminant structure of the data space, RLLDCF constructs the local predictors and derives a local regression function. Briefly, the local regression function is mainly to obtain the relation by modeling $X_i$ to the new representation $V_i$, which is formulated as

$$\min_{G_i, b_i} \frac{1}{n_i}\|V_i - G_i^T X_i - b_i \mathbf{1}_{n_i}^T\|_F^2 + \lambda_{RLLDCF}\|G_i\|_F^2, \quad (39)$$

where $\lambda_{RLLDCF}$ is a positive parameter, $n_i$ is the number of data points in each local region of $X_i$. That is, RLLDCF jointly solves problem in Eq.(39) and the matrix decomposition task, so that the learnt new representation can contain much discriminant structure information, so that the representation learning performance can be potentially improved.

**SDCF$_2$ [128].** To solve the problem that the distant repulsion property of data is usually neglected resulting in distorted embedding maps, SDCF$_2$ includes the distant repulsion constraints into CF. Specifically, the dissimilar data points in original high-dimensional data space can be kept far apart in the learnt low-dimensional feature space, which leads to the following distant repulsion constraint for SDCF$_2$:

$$\min_V \frac{1}{2}\sum_{i=1}^{N}\sum_{j=1}^{N} \Upsilon_{ij} \exp\left(-\|V_i - V_j\|^2\right), \quad (40)$$

where $\Upsilon_{ij} = \|X_i - X_j\|^2$ is the repulsive weight. Note that Eq. (40) is added into CF to make full use of the repulsive property of data, which can clearly force the dissimilar samples to be far away in the low-dimensional representation space.

## 5. Unspervised Multi-view CF Methods

In real-world emerging applications, multi-view data can be encountered, since real data may have different representations in multiple views or come from different sources [194-196]. For example, Zhang et al. [194] proposed the first binary multi-view clustering algorithm, which provides a feasible solution to the challenging large-scale data clustering problem in an efficient manner. While most existing CF-based methods perform matrix decomposition and clustering only relying on individual view, which has obvious limitations when dealing with multi-view datasets. To overcome this problem, both MVCC [122] and MVCF [123] explore extending the traditional individual-view CF methods to the multi-view scenario. Note that the general multi-view CF framework has been presented in Fig.9.

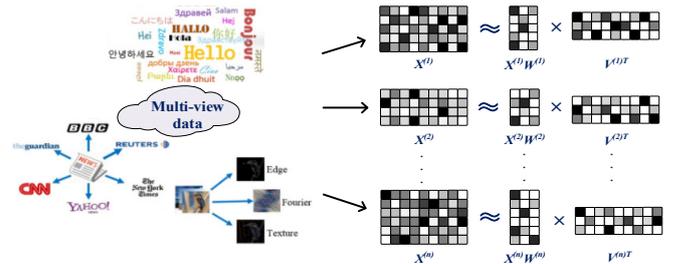

**Fig.9:** Multi-view CF method.

**MVCC [122].** MVCC jointly incorporates the CF, the local manifold regularization and the consistency constraint into a unified framework. MVCC differs from the other multi-view learning methods that treat each view equally, MVCC assigns different weights for each view and then drives a consensus solution across all the views. For a given $n_\varpi$-view dataset $X = \left[X^{(1)}, X^{(2)}, ..., X^{(n_\varpi)}\right]$, where $X^\varpi \in \mathbb{R}^{D_\varpi \times N}$ denotes the samples of the $\varpi$-th view and $D_\varpi$ is the dimension of the $\varpi$-th view, then MVCC minimizes the following reconstruction error:

$$\sum_{\varpi=1}^{n_\varpi} g\left(W^{(\varpi)}, V^{(\varpi)}\right) = \left\| X^{(\varpi)} - X^{(\varpi)} W^{(\varpi)} \left(V^{(\varpi)}\right)^T \right\|_F^2, s.t. W^{(\varpi)}, V^{(\varpi)} \geq 0. \quad (41)$$

Then, MVCC incorpoartes the local manifold regularization for each view similarly as LCCF using heart kernel weights and minimizes the multi-view regularization $tr\left(\left(V^{(\varpi)}\right)^T L^{(\varpi)} V^{(\varpi)}\right)$. To further explore the relationship among all views and derive the common consensus representation matrix, MVCC clearly minimizes the consistency loss punishment function $\left\| V^\varpi - V^* \right\|_F^2$, where $V^*$ is the common consensus matrix. Finally, MVCC solves the following joint minimization problem:

$$\sum_{\varpi=1}^{n_\varpi} g\left(W^{(\varpi)}, V^{(\varpi)}\right) + \alpha_{MVCC} tr\left(\left(V^{(\varpi)}\right)^T L^{(\varpi)} V^{(\varpi)}\right) + \beta_{MVCC} \delta_\varpi \left\| V^\varpi - V^* \right\|_F^2,$$
$$s.t. W^{(\varpi)}, V^{(\varpi)}, \delta_\varpi \geq 0, \sum \delta_\varpi = 1 \quad (42)$$

where $\alpha_{MVCC}$ and $\beta_{MVCC}$ are two trade-off parameters and $\delta_\varpi$ denotes the weights of the $\varpi$-th view among all views.

**MVCF [123].** MVCF is also a multi-view clustering method. The main difference between both MVCC and MVCF is that MVCC applies the heat kernel weighting and pre-computes the weights, while MVCF adopts an adaptive weighting scheme like CFANs [12]. In detail, MVCF explots the local geometry of data distribution by optimizing the graph matrix in a global view. Finally, MVCF minimizes the following criterion:

$$\sum_{\pi=1}^{n_\pi} \omega_\pi \left\{ g\left(W^{(\pi)}, V^{(\pi)}\right) + \sum_{i,j=1}^N \left\| \left(v_i^\pi\right)^T - \left(v_j^\pi\right)^T \right\|_2^2 \left(S_{ij}^{AN}\right)^\lambda \right\} + \alpha_{MVCF} \sum_{\pi=1}^{n_\pi} (\omega_\pi)^2$$
$$s.t. W^{(\pi)}, V^{(\pi)}, \omega_\pi, S_{ij}^{AN} \geq 0, \sum \omega_\pi = 1, \forall_j \mathbf{1}^T S_j^{AN} = 1 \quad (43)$$

where $\alpha_{MVCF}$ is the trade-off parameter. For the parameter $\omega_\pi$ tuning the weight of the $\pi$-th view, MVCF adds an extra term $\sum_{\pi=1}^{n_\pi} (\omega_\pi)^2$ to avoid the trivial solution, i.e., only the weight of one view is learned as 1 and the others are 0.

### B. Supervised CF variants

Unsupervised CF variants mainly focus on learning representations, but they cannot utilize any supervised prior information of samples, such as class label information, even though the labels of samples are available. Therefore, to make full use of the available label information and further improve the discriminative abilities of the obtained feature representations, supervised CF methods have also been explored, which can be further divided into fully-supervised ones and semi-supervised ones.

**SGDCF [22].** SGDCF is a classical supervised CF method. For the training data $X \in \mathbb{R}^{D \times N}$ with labels, SGDCF first defines a class indicator matrix $C^{SGDCF} \in \mathbb{R}^{c \times N}$ using label information:

$$C_{ij}^{SGDCF} = \begin{cases} 1, & \text{if } label(j) = i, j=1,2,...,N, i=1,...,c \\ 0, & \text{otherwise} \end{cases} \quad (44)$$

where $label(j)$ is the class label of $x_j$. Then, the label constraint term can be defined as $\left\| C - \Psi V^T \right\|_F^2$, where $\Psi$ denotes a nonnegative auxiliary matrix which can be initialized randomly. In addition to using label information, SGDCF also unifies the graph regularization in terms of $\left\| V^T - LV^T \right\|_F^2$ so that the local geometry structure can be retained, where the graph Laplacian

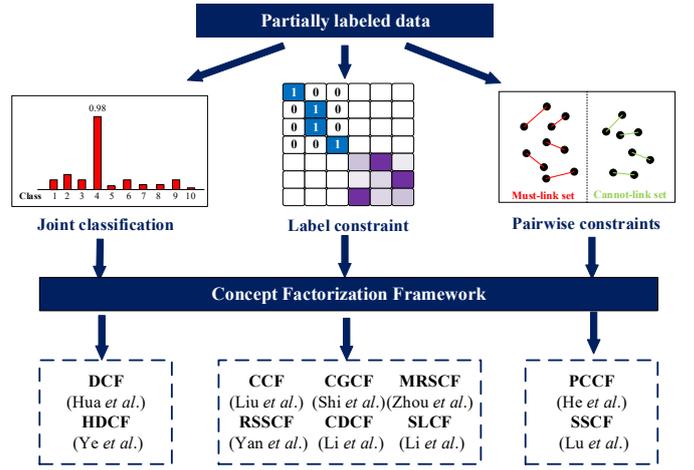

Fig.10: Semi-supervised learning strategies for CF.

matrix $L = D_S - S$ and SGDCF defines the weight matrix $S$ by using the '0-1' weighting approach.

By combining the concept factorization term, the graph regularization term and the label constraint term, we can obtain the objective function of SGDCF as follows:

$$O_{SGDCF} = \left\| X - XWV^T \right\|_F^2 + \alpha_{SGDCF} \left\| C^{SGDCF} - \Psi V^T \right\|_F^2 + \beta_{SGDCF} tr\left(V^T L V\right), s.t. W, V, \Psi \geq 0 \quad (45)$$

where $\alpha_{SGDCF}$ and $\beta_{SGDCF}$ are nonnegative trade-off parameters.

### C. Semi-supervised CF variants

Based on using all the labeled data, fully-supervised CF methods have significantly enhanced the representation ability, but in real applications, it is usually hard and costly to obtain the labels of data. As a result, the applications of fully-supervised methods may be restricted. Under this circumstance, the semi-supervised learning methods that can utilize a small number of labeled data and a large amount of unlabeled data tend to outperform the fully-supervised methods in terms of performance and application scenario [156-157]. As such, researchers also investigated the semi-supervised CF-based methods to enhance quality and ability of the representation learning.

As shown in Fig.10, there are three popular strategies to extend CF to semi-supervised scenario, i.e., adding the label constraint into factorization, joint representation learning and classification, and adding pairwise constraints into CF framework. Next, we will introduce the three strategies in detail.

### 1. Joint Classification based CF Methods

We first introduce the unified CF frameworks that aims to improve the representation learning ability through performing joint representation and classification. Several classical methods of this kind are described as Fig.10.

**DCF [23].** DCF is the first work of extending CF to semi-supervised scenario, which combines the representation learning with the task of classification. To make full use of the partial labeled data, DCF jointly optimizes the representation issue and trains a classifier. In particular, for a given partial-labeled data matrix $X = [X_L, X_U] \in \mathbb{R}^{D \times (l+u)}$, where $X_L$ denotes a labeled set and $X_U$ is an unlabeled set, $l$ and $u$ are the numbers of labeled and unlabeled samples respectively. Assume that the label set of $X_L$ is $Y = [y_1, y_2, ..., y_l] \in \mathbb{R}^{c \times l}$, DCF trains a linear function $\mathbf{f}(v) = B_{DCF} v$ for classification from sample-label pairs $(v_i^T, y_i)$, where $v_i^T$ is the new representation of $x_i$, $i \in \{1, 2, ..., l\}$, $B_{DCF}$ is a $c \times r$ coefficient matrix. The procedure of training this function $\mathbf{f}(v)$ can be mathematically formulated as follows:

$$\min_{B_{DCF}} \sum_{i=1}^{l} \|y_i - B_{DCF} v_i^T\|_F^2 + \gamma_{DCF} \|B_{DCF}\|_F^2, \quad (46)$$

where $\gamma_{DCF}$ is a nonnegative parameter and $\|B_{DCF}\|_F^2$ is added to avoid the overfitting. Using the obtained classifier $\mathbf{f}(v)$, DCF can compute $y_i = \mathbf{f}(v_i)$ and assign class label for each unlabeled data point $x_i$ as $\arg\max_j y_{ij}$. Since $Y$ is a $c \times l$ matrix and $V^T$ is a $r \times (l+u)$ matrix, to make a relationship between both $Y$ and $V^T$, DCF defines a selection matrix $C_{DCF} = [e_1, e_2, ..., e_l] \in \mathbb{R}^{N \times l}$, where $e_i, i \in \{1, 2, ..., l\}$ denotes an $N$-dimensional vector with the $i$-th entry being 1 and the other entries being 0. Finally, the minimization based objective function of DCF is defined as

$$O_{DCF} = \|X - XWV^T\|_F^2 + \alpha_{DCF} \left( \|Y - B_{DCF} V^T C_{DCF}^T\|_F^2 + \gamma_{DCF} \|B_{DCF}\|_F^2 \right). \quad (47)$$

**HDCF [24].** Although DCF can obtain the discriminative representations, it still fails to uncover the intrinsic geometrical structure of data. To obtain both the discriminative and locality-preserving representations, HDCF incorporates hyper-graph regularizer into DCF, which can clearly preserve the geometrical structures. Note that the strategy of making full use of the prior label information is the same for both HDCF and DCF.

## 2. Label Constraint based CF Methods

Different from the joint classification based CF methods, label constraint based CF methods utilize the prior label information of the labeled data as an additional constraint to guide the semi-supervised factorization. Several representative label constraint based CF methods are described as follows:

**CCF [17].** CCF is the most representative method of this kind. To improve the discriminating power of learnt representations, CCF clearly extends CF to semi-supervised scenario and guides the constrained CF by defining a label constraint matrix $A$. Let $A_L \in \mathbb{R}^{l \times c}$ be the class indicator matrix defined on labeled data $X_L$. The element $(A_L)_{ij}$ is defined as 1 if $x_i$ is labeled as the $j$-th class, and 0 otherwise. Note that CCF did not define an explicit class indicator for $X_U$ and simply used an identity matrix $I_{u \times u}$ of dimension $u \times u$ for $X_U$. Thus, CCF defines the overall label constraint matrix $A$ as follows:

$$A = \begin{bmatrix} (A_L)_{l \times c} & 0 \\ 0 & I_{u \times u} \end{bmatrix} \in \mathbb{R}^{(l+u) \times (c+u)}. \quad (48)$$

To ensure the samples sharing the same label can be mapped into the same class in low-dimensional space (i.e., same $v_i$) [17], CCF imposes the label constraint by an auxiliary matrix $Z$:

$$V = AZ. \quad (49)$$

By substituting $V = AZ$ into CF, CCF computes a nonnegative matrix $W \in \mathbb{R}^{N \times R}$ and an auxiliary matrix $Z \in \mathbb{R}^{(c+u) \times r}$ from

$$O_{CCF} = \|X - XWZ^T A^T\|_F^2, \text{ s.t. } W, Z \geq 0. \quad (50)$$

**Algorithm 3: CCF algorithm**
**Input:** Partial labeled data $X = [X_L, X_U] \in \mathbb{R}^{D \times N}$, rank $r$ and a small constant $\varepsilon$.
**Initialization:** Construct the kernel matrix $K = X^T X$; Initialize $W$ and $V$ to be random matrices; Construct the label constraint matrix $A$ by Eq.(46).
*While not converged do*
1. Update $w_{ik}^{t+1}$ by $w_{ik}^t (KAZ)_{ik} / (KWZ^T A^T AZ)_{ik}$;
2. Update $z_{jk}^{t+1}$ by $z_{jk}^t (A^T KW)_{jk} / (A^T AZW^T KW)_{jk}$;
3. Convergence check: if $\|(O_{CCF})^{t+1} - (O_{CCF})^t\| \leq \varepsilon$, stop; else, return to step 1.
*End while*
**Output:** New low-dimensional representation $V$ of $X$.

The optimization procedure of CCF is shown in Algorithm 3.

Although CCF can obtain the discriminative representations using the label constraints, it still has several drawbacks: 1) it may be sensitive to noise and outliers by performing representation learning directly in original data space and using the Frobenius norm to encode the reconstruction error; 2) it fails to preserve the local manifold information during decomposition; 3) it aims to map the intra-class data points into the same concept, but note that this is infeasible if there is only one labeled sample to rely on; 4) it simply defines the label indicator for the unlabeled samples as an identity matrix, that is, CCF may not fully utilize the unlabeled samples and moreover fails to predict the class label for unlabeled samples. As such, several subsequent studies have been done to address these issues.

**RSSCF [31].** To inherit the merits of CCF and further improve the robustness properties to noise and outliers of CCF, RSSCF has been recently proposed. For the consideration of robustness, different from CCF that uses the Frobenius norm to measure the reconstruction error, RSSCF replaces it by $L_{2,1}$-norm due to the fact that $L_{2,1}$-norm was proven to be robust to noise and outliers [152-153]. Besides, to encode the reconstruction error jointly, RSSCF minimizes a $L_{2,1}$-norm based reconstruction error $\|X - XWZ^T A^T\|_{2,1}$ and $L_{2,1}$-norm is also regularized on $Z$, i.e., $\|Z\|_{2,1}$, to obtain the sparse representation. Overall, RSSCF solves the following minimization problem as

$$O_{RSSCF} = \|X - XWZ^T A^T\|_{2,1} + \alpha_{RSSCF} \|Z\|_{2,1}, \text{ s.t. } W, Z \geq 0, \quad (51)$$

where $\alpha_{RSSCF}$ denotes a nonnegative trade-off parameter.

**CGCF [129], SLCF [137], LRCCF [30], CCF-GL [32] and HCCF [132].** These extensions are proposed to compute the locality-preserved discriminative representations by using the label constraints and considering the geometrical structures. Specifically, CGCF combines CCF and LCCF into a unified model to enhance the locality of CCF. SLCF incorporates the idea of LCF into CCF. LRCCF includes a local regularization constraint based on the representation $Z^T A^T$. CCF-GL not only incorporates the graph Laplacian over $Z^T A^T$, but also utilizes the label information of data by building the cannot-link pairwise constraints on $Z^T A^T$, where the graph Laplacian is able to preserve the geometrical information of data and the cannot-link pairwise constraints can impose the restrictions on inter-class data points. It is worth noting that HCCF differs from CGCF, SLCF, LRCCF and CCF-GL that it extends CCF to the hyper-graph scenario, i.e., it incorporates the hyper-graph regularization into CCF so that it can better handle more complex data distribution with multivariate relationships.

**CDCF [34].** CDCF addresses the drawback that CCF cannot process the case that there is only one labeled data point to train in each class. Specifically, CDCF associates the class labels of samples with their representations by introducing a class-driven constraint as $\sum_j d_{y_j}^T v_j = tr(\mathcal{D}V^T)$, where $y_i$ is the label of sample $x_i$ and $\mathcal{D}=[d_{y_1}, d_{y_2},...,d_{y_N}]^T$ is the indicator matrix for the inhomogeneous representation. Therefore, CDCF has the potential to force the representations of data points to be more similar within one class, but different between classes.

**RS²ACF [33].** Inspired by the idea of label propagation for semi-supervised learning [147-152], RS²ACF includes the label prediction into CCF. Different from CCF that simply defines the label constraint sub-matrix for unlabeled data by an identity matrix, RS²ACF explicitly learns a label indicator for $X_U$ so that it can also make sure the unlabeled data with the same predicted labels to be mapped to be close in feature space. To be specific, RS²ACF designs the label constraint matrix as

$$A = \begin{bmatrix} (A_L)_{l \times c} & 0 \\ 0 & (A_U)_{u \times c} \end{bmatrix} \in \mathbb{R}^{(l+u) \times 2c}, \quad (52)$$

where $A_U$ is the class indicator for unlabeled data $X_U$, which is learned by solving a label predictor $P \in \mathbb{R}^{D \times c}$ from:

$$\min_{A_U,P} \|A_L - X_L^T P\|_F^2 + \|A_U - X_U^T P\|_F^2 + \|P\|_{2,1}, \quad (53)$$

from which one see that RS²ACF clearly propagates the label information from $X_L$ to $X_U$ by $P$. Note that to make the predicted results reasonable, $P$ is initialized as a label predictor explicitly based on the labeled data [33]. Furthermore, RS²ACF also involves the adaptive weight learning to retain the manifold structures of the original data space, new representation space and the label space at the same time.

**SDCF₁ [44] and GDCF [44].** SDCF₁ and GDCF provide another way for doing semi-supervised CF. Specifically, SDCF₁ expects that the classes of dataset can be placed in a clear separated cluster in the resulting representation space $V$. The label matrix $\Omega \in \mathbb{R}^{c \times N}$ of SDCF₁ is defined as follows:

$$\Omega_{ij} = \begin{cases} 1 & \text{if sample } x_j \text{ is labeled as class } i \\ 0 & \text{otherwise} \end{cases}. \quad (54)$$

With the constructed $\Omega$, the label constraint is defined as $\|\Omega - \Pi \hat{V}^T\|^2$, where $\hat{V}=[v_1,...,v_l,0,...,0]^T \in \mathbb{R}^{N \times r}$ and the label matrix $\Pi \in \mathbb{R}^{c \times r}$. The label matrix $\Pi$ can linearly transform and scale the vectors in $V$ to best fit the label matrix $\Omega$ [44]. Finally, the objective function of SDCF₁ is formulated as

$$O_{SDCF} = \|X - XWV^T\|_F^2 + \alpha_{SDCF} \|\Omega - \Pi \hat{V}^T\|^2, \text{ s.t. } W, V \geq 0. \quad (55)$$

Note that SDCF₁ fails to extract the latent concepts consistent with the manifold geometry, so GDCF incorporates the LCCF-style graph Laplacian regularizer into the SDCF₁ model to preserve the local manifold structures of data.

**MCC-Based Robust Semi-supervised CF (MRSCF) [131].** Motivated by [141], MRSCF utilizes the label information by the following minimization term:

$$tr\left[(V^T - \Omega)\mathcal{H}(V^T - \Omega)^T\right], \quad (56)$$

where $\Omega$ is the indicator matrix of size $c \times N$, which is defined in Eq.(54). $\mathcal{H}$ is a diagonal matrix with $\mathcal{H}_{ii}=1$ if $x_i$ is labeled and $\mathcal{H}_{ii}=0$ otherwise. Note that MRSCF includes this term to measure the distance between the representation and indicator matrices of the labeled samples and encourage them close to each other in the representation space.

**Remarks.** Incorporating the label information as a hard constraint is the most common way to extend CF to the supervised or semi-supervised cases. However, the hard label information is always hard to obtain in reality. Thus, to fully utilize the small number of labels, pairwise constraints (PCs) based semi-supervised CF methods have also been developed, since PCs can be achieved with minimal human efforts and can provide more supervision information compared with the traditional label information. Moreover, PCs are more flexible in regulating the supervised information than the hard label information.

### 3. Pairwise Constraints based CF Methods

It is noted that the PCs also offer advantages over the traditional label information, since the supervised prior knowledge can be enriched by constructing the pairwise constraints, i.e., must-link (ML) and cannot-link (CL) constraints, based on the limited label information [86-87]. In addition, the label constrained semi-supervised CF methods also neglect the intra-class variance, which may lead to the decreased results. Next, we introduce the representative pairwise constraints based CF methods:

**PCCF [25].** To obtain discriminative representations, PCCF encourages the samples under pairwise ML constraints should have the same label and the samples under CL constraints to have different labels as much as possible. PCCF firstly defines a ML constraints symmetric matrix $\mathcal{M}=[m_{ij}] \in \mathbb{R}^{N \times N}$ and a CL constraints symmetric matrix $\mathcal{C}=[c_{ij}] \in \mathbb{R}^{N \times N}$ as follows:

$$m_{ij} = \begin{cases} 1 & \text{if } x_i, x_j (i \neq j) \text{ have the same class label} \\ 0 & \text{otherwise} \end{cases}, \quad (57)$$

$$c_{ij} = \begin{cases} 1 & \text{if } x_i, x_j (i \neq j) \text{ have different class label} \\ 0 & \text{otherwise} \end{cases}. \quad (58)$$

After combining the cost function of CF, the cost function for the violation of pairwise ML and CL constraints, PCCF solves the following minimization problem:

$$O_{PCCF} = \|X - XWV^T\|_F^2 + \alpha_{PCCF} \sum_{j=1}^{N} \left( \sum_{i:m_{ij}=1} \sum_{c=1}^{r} \sum_{h=1, h \neq c}^{r} v_{jc} v_{ih} + \sum_{i:c_{ij}=1} \sum_{c=1}^{r} v_{jc} v_{qc} \right), \quad (59)$$

where $\alpha_{PCCF} \geq 0$ denotes a non-negative trade-off parameter.

**SSCF [26].** Different from PCCF that imposes a constant penalty for all pairwise constraints, an optimized method called SSCF provides a dynamic penalty mechanism. More specifically, SSCF mainly aims at allowing the dissimilar data points of the same label to be mapped farther than the similar ones so that the intra-class variance can be well accounted.

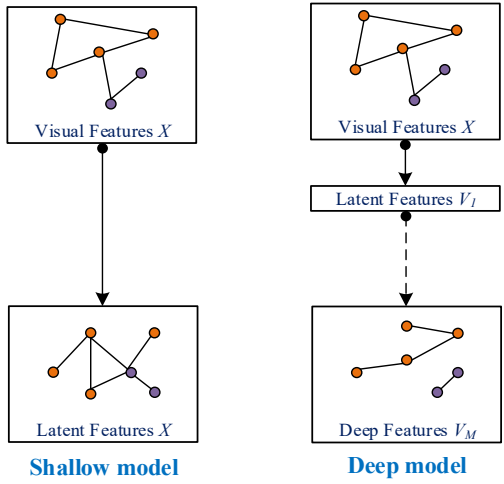

**Fig.11:** Comparison of shallow and deep CF models.

**CNPCF [27] and RCF [28].** Although PCCF can compute the discriminant representations using the pairwise constraints, it still has a shortcoming that it only utilizes prior knowledge but neglecting the proximity information of the whole distribution in datasets. By handling this issue, CNPCF also considers retaining the local structures of whole dataset to guide clustering, in addition to using supervised prior knowledge. Note that CNPCF utilizes the prior information by the ML constraints to modify the graph. But if the prior knowledge is few, only using this information to modify the graph is in fact not effective to preserve the geometric structure of data. To solve this issue, a regularized CF (RCF) [28] firstly propagates the limited dual connected constraints to whole dataset, and then constructs a novel weight matrix which can be seen as a penalty term used to make the intra-class data points more compact and the inter-class sample more separable in the feature space.

## IV. DEEP/MULT-LAYER CF-BASED METHODS

By considering the characteristics of locality, robustness, operating space, discriminability and multiple-views, existing unsupervised CF variants have obtained obvious performance improvement over standard CF method, but they still suffer from a serious drawback, i.e., they cannot uncover hidden deep features from complex real data. It should be noted that in recent years deep learning-based methods have achieved great success in the area of visual representation learning, speech recognition and natural language processing, therefore the related researchers have also paid a lot of attention to exploring how to extend the traditional shallow (or single-layer) CF models to the deep (or multi-layer) scenario. As illustrated in Fig.11, the shallow model directly maps the visual features into latent space while deep model can learn the hierarchical feature representations from visual features. In this paper, we will divide the existing deep CF methods into two major categories, i.e., traditional deep CF models and optimized deep CF models.

### A. Traditional Feeding-style Deep CF Models

The deep CF models of this kind usually performs the standard CF process to obtain the intermediate new representation, and

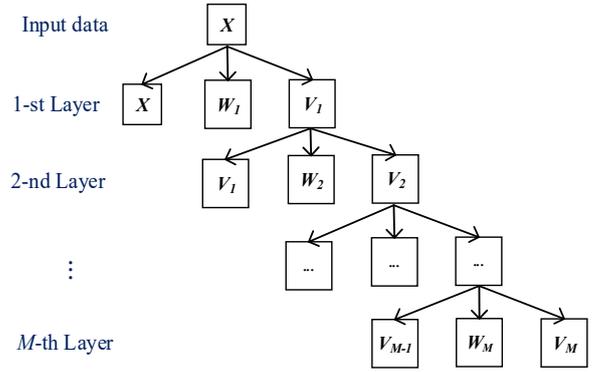

**Fig.12:** The simple factorization structure of the traditional multi-stage deep CF models.

---

**Algorithm 4: MCF algorithm**
**Input:** Data matrix $X \in \mathbb{R}^{D \times N}$, a fixed number $M$ of layers, rank $r$ and a small constant $\varepsilon$.
**Initialization:** Initialize $X_1 = X$ in the first layer; Initialize $W$ and $V$ to be random matrices.
For $m=1$ to $M$ do
  Construct the kernel matrix $K = X_m^T X_m$;
  *While not converged do*
  1. Update $(w_{ik}^{t+1})_m$ by $(w_{ik}^t)_m ((KV)_{ik})_m / ((KWV^TV)_{ik})_m$;
  2. Update $(v_{jk}^{t+1})_m$ by $(v_{jk}^t)_m ((KW)_{jk})_m / ((VW^T KW)_{jk})_m$;
  3. Convergence check: if $\|(O_{CCF})^{t+1} - (O_{CCF})^t\| \le \varepsilon$, stop; else, return to step 1.
  *End while*
  $X_{m+1} = V_m^T$;
End
**Output:** New low-dimensional representation $V^*$ of $X$.

---

then feed the intermediate representation of the previous layer directly as the input of the next layer for further decomposition, as shown in Fig.12. That is, these traditional deep models did not consider how to solve the representation by the optimized strategies. Classical traditional deep CF models include:

**MCF [35].** The first work that extends the standard CF model to deep scenario is called MCF [35], proposed in 2015. Inspired by the multilayer NMF method [39], MCF adopts a kind of simple deep hierarchy and performs the decomposition sequentially. Suppose that the hierarchical structure has $M$ layers, MCF performs the matrix decomposition as CF in the first layer, that is, $X = X_1 W_1 V_1^T$. Then MCF considers the new representation learnt in the first layer, i.e., $V_1^T$, as the input of the second layer, and performs the similar factorization as $V_1^T = V_1^T W_2 V_2^T$. Similarly, in the following layers, MCF directly uses the low-dimensional representation of the previous layer as the input of the next layer, which is formulated as follows:

$$X = XW_1 V_1^T, \ V_1^T = V_1^T W_2 V_2^T, ..., \ V_{M-1}^T = V_{M-1}^T W_M V_M^T, \quad (60)$$

i.e., $X = XW_1 V_1^T W_2 V_2^T ... W_M V_M^T$, $W = W_1 V_1^T W_2 V_2^T ... W_M$ and $V^T = V_M^T$. The simple hierarchical structure of this kind is shown in Fig.12, and the objective function of MCF is defined as

$$O_{MCF} = \left\| X_m - X_m W_m V_m^T \right\|_F^2, \quad s.t. \quad W_m, V_m \geq 0, \quad m = 1, 2, ..., M, \quad (61)$$

where $X_m W_m$ and $V_m^T$ denote the basis vectors and the learnt new representations in the $m$-th layer, respectively. We summarize the optimization procedure of MCF in Algorithm 4.

**GMCF [36].** Although MCF can obtain the deep hierarchical information by the designed deep structure, it fails to retain the local manifold structure of samples. As such, GMCF [36] was then proposed to improve MCF by integrating the geometrically-based graph regularization into MCF to discover the local geometrical information of data, which can also be seen as a deep version of LCCF. Note that the deep factorization principle of GMCF is similar in sprit to that of MCF. However, in the $m$-th layer, GMCF needs to construct a $p$-nearest neighbor graph over input data $X_m = \left[ (x_1)_m, (x_2)_m, ..., (x_N)_m \right]$, then define a weight matrix $S_m$ using the cosine similarity, and finally obtain the graph Laplacian as $L_m = D_m - S_m$, where $D_m$ is a diagonal matrix with $(D_m)_{ii} = \sum_j (S_m)_{ij}$. The objective function of GMCF is defined as

$$O_{GMCF} = \left\| X_m - X_m W_m V_m^T \right\|_F^2 + \alpha_{GMCF} tr\left( V_m^T L_m V_m \right), \quad (62)$$
$$s.t. \quad W_m, V_m \geq 0, \quad m = 1, 2, ..., M$$

where $\alpha_{GMCF}$ is nonnegative parameter to trade-off the reconstruction error and local manifold preservation.

**DGMCF [164].** Since GMCF only considers preserving the locality information in data space, DGMCF improves it by considering keep locality information both in data and feature space at the same time. In other words, DGMCF is can be regarded as a deep extension of GCF [8]. In each layer, it constructs $p$-nearest neighborhood data and feature graphs using binary weights. Finally, it solves the following joint objective function:

$$O_{DGMCF} = \left\| X_m - X_m W_m V_m^T \right\|_F^2 + \alpha_{DGMCF} tr\left( V_m^T L_m^V V_m \right)$$
$$+ \beta_{DGMCF} tr\left( W_m^T L_m^W W_m \right), \quad (63)$$
$$s.t. \quad W_m, V_m \geq 0, \quad m = 1, 2, ..., M$$

where $\alpha_{DGMCF}$ and $\beta_{DGMCF}$ are two nonnegative parameters, $L_m^V$, $L_m^W$ are the graph Laplacian matrices of data graph and feature graph respectively in the $m$-th layer.

**Remarks.** The deep factorization models of MCF, GMCF and DGMCF provide a simple feeding-style way to extend single-layer CF to deep versions. This simple factorization structure in Fig.12 can also be extended for the other shallow CF methods to deep methods. Thus, we summarize the simple deep methods in this category as traditional feeding-style deep CF models. However, it should be noted that the simple "feeding" strategy may be ineffective and even unreasonable in reality, since we cannot strictly ensure that the previous representation is optimal and also effective for subsequent layers. Note that the learned representations in the first layer may be inaccurate and may lose important feature information, which will cause the reconstruction errors to be larger and larger with the increase of layers.

*B. Optimized Deep CF Models*

Toward the shortcomings of the traditional multi-stage deep CF models, researchers also investigated to refine the deep network models to discover the hidden deep information by optimizing

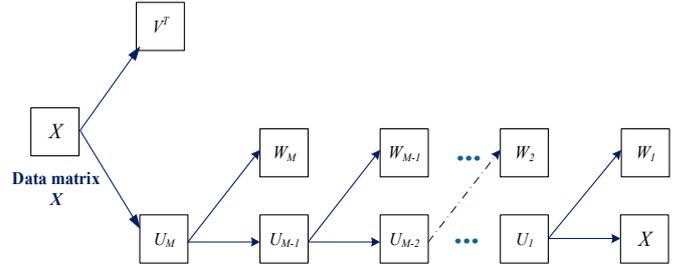

**Fig.13:** The deep factorization structure of DSCF-Net.

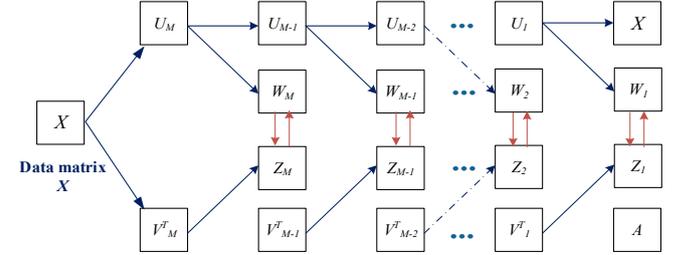

**Fig.14:** The deep factorization structure of DS²CF-Net.

the basis concepts or new representations in each layer, rather than using the simple "feeding" strategy. Note that two classical optimized deep matrix factorization models are called Deep Semi-NMF (DSNMF) [40] and Weakly-supervised Deep MF (WDMF) [41]. Both DSNMF and WDMF have provided new ways to optimize the deep factorization models, which can uncover the hidden deep features using multiple layers of linear transformations and updating the basis concepts or new representations in each layer. But note that DSNMF is a two-stage method: in the first stage, DSNMF adopts the traditional "feeding" deep factorization strategy as MCF; in the second stage, it refines the new representation and basis vectors directly based on the outputs of each layer in the first stage by an independent step. As a result, DSNMF will also suffer from the same performance-degrading issue as the above traditional multi-stage deep CF models. In addition, DSNMF and WDMF cannot encode the local geometry structure of the learned representations in each layer, especially in an adaptive manner. Moreover, they directly decompose data in the original space that usually has noise and corruptions, which may decrease the performance. In more recent years, more optimized strategies have been proposed to address the above drawbacks, which will be introduced below:

**DSCF-Net [37].** DSCF-Net proposed in 2019 designs a novel deep network structure for deep CF, which is illustrated in Fig.8, where $\|E\|_{2,1}$ is the sparse error by $L_{2,1}$-norm regularization and $X-E$ is the noise-removed clean data. To improve the robustness to noise, DSCF-Net incorporates the subspace recovery process into the process of CF and factorizes data in the recovered clean space. Specifically, DSCF-Net factorizes the recovered data matrix $X - E$ into $M+1$ factors, i.e., $V^T, U_M, ..., U_1$, where the output of the first layer is transformed from the visual space, i.e., $U_1 = W_1(X - E)$. To discover deep hidden information, DSCF-Net uses $M$ layers of linear transformations and the mathematical representation of the deep structure is formulated as follows [37], which is also illustrated in Fig.13.

**Table II.** Overall summarization of existing CF-based representation learning methods.

| Categories of current CF-based methods | | | Methods | Properties | | | | Publication |
|---|---|---|---|---|---|---|---|---|
| | | | | *Local* | *Robust* | *Kernelized* | *Discriminant* | |
| **Single-layer CF methods** | Unsupervised CF | Locality-preserving CF | Graph Regularized CF | LCCF | √ | | | | Cai *et al.* 2011 [7] |
| | | | | GCF | √ | | | | Ye *et al.* 2014 [8] |
| | | | | RDCF | √ | √ | | | Guo *et al.* 2015 [121] |
| | | | | HRCF | √ | | | | Li *et al.* 2015 [133] |
| | | | | SMCF | √ | | | | Li *et al.* 2016 [135] |
| | | | | LGCF$_1$ | √ | | | | Li *et al.* 2017 [10] |
| | | | | LGCF$_2$ | √ | | | | Qian *et al.* 2017 [139] |
| | | | | GRLCF | √ | | | | Ye *et al.* 2017 [9] |
| | | | | GRLCF$_{CLR}$ | √ | | | | Ye *et al.* 2017 [9] |
| | | | | DHCF | √ | | | | Ye *et al.* 2017 [134] |
| | | | | ADGCF | √ | √ | | | Ye *et al.* 2017 [13] |
| | | | | HDCF | √ | | | | Ye *et al.* 2018 [24] |
| | | | | RLLDCF | √ | √ | | √ | Jiang *et al.* 2018 [125] |
| | | | | MCFAW | √ | | | | Shu *et al.* 2018 [11] |
| | | | | CFANs | √ | | | | Pei *et al.* 2018 [12] |
| | | | | CF-OGL | √ | | | | Shu *et al.* 2019 [126] |
| | | | | RFA-LCF | √ | √ | | | Zhang *et al.* 2019 [14] |
| | | | Local coordinate coding based CF | LCF | √ | | | | Liu *et al.* 2011 [16] |
| | | | | GLCF | √ | | √ | | Li *et al.* 2015 [18] |
| | | | | LGCF$_1$ | √ | | | | Li *et al.* 2017 [10] |
| | | | | GRLCF | √ | | | | Ye *et al.* 2017 [9] |
| | | | | GRLCF$_{CLR}$ | √ | | | | Ye *et al.* 2017 [9] |
| | | | | RFA-LCF | √ | √ | | | Zhang *et al.* 2019 [14] |
| | | | Self-representation based CF | SRMCF | √ | | | | Ma *et al.* 2018 [20] |
| | | | | JSGCF | √ | | | | Peng *et al.* 2019 [21] |
| | | | Other locality-preserving ways for CF | SDRCF | √ | | | | Du *et al.* 2013 [120] |
| | | | | LRCF | √ | | √ | | Shu *et al.* 2015 [30] |
| | | | | SCF | √ | | | | Shen *et al.* 2020 [127] |
| | | | | RSSCF | √ | | | | Shen *et al.* 2020 [127] |
| | | | | LSDCL | √ | | | | Zhao *et al.* 2020 [183] |
| | | | | CFLCs | √ | | | | Chen et al. 2020 [186] |
| | | Kernel CF | | MKCF | √ | | √ | | Li *et al.* 2012 [19] |
| | | | | GLCF | √ | | √ | | Li *et al.* 2015 [18] |
| | | | | GMKCF | | | √ | | Li *et al.* 2019 [136] |
| | | | | DMKCF | √ | | √ | √ | Mu *et al.* 2020 [182] |
| | | Robust CF | | RDCF | √ | √ | | √ | Guo *et al.* 2015 [121] |
| | | | | ADGCF | √ | √ | | | Ye *et al.* 2017 [13] |
| | | | | GCCF | √ | √ | | | Peng *et al.* 2018 [124] |
| | | | | RLLDCF | √ | √ | | √ | Jiang *et al.* 2018 [125] |
| | | | | RSCF | √ | √ | | | Shen *et al.* 2020 [127] |
| | | Discriminant CF | | RDCF | √ | √ | | | Guo *et al.* 2015 [121] |
| | | | | SDCF$_2$ | √ | | | √ | Shu *et al.* 2017 [128] |
| | | | | RLLDCF | √ | √ | | √ | Jiang *et al.* 2018 [125] |
| | | Multi-view CF | | MVCC | √ | | | | Wang *et al.* 2016 [122] |
| | | | | MVCF | √ | | | | Zhan *et al.* 2018 [123] |
| | Semi-supervised CF | Joint classification based CF | | DCF | | | | √ | Hua *et al.* 2011 [23] |
| | | | | HDCF | √ | | | √ | Ye *et al.* 2018 [24] |
| | | Label constraint based CF | | CCF | | | | √ | Liu *et al.* 2014 [29] |
| | | | | CGCF | √ | | | √ | Shi *et al.* 2014 [129] |
| | | | | HCCF | | | | √ | Li *et al.* 2015 [132] |
| | | | | CCF-GL | √ | | | √ | Lu *et al.* 2016 [32] |
| | | | | CDCF | | | | √ | Li *et al.* 2016 [34] |
| | | | | SDCF$_1$ | | | | √ | Li *et al.* 2016 [44] |
| | | | | GDCF | √ | | | √ | Li *et al.* 2016 [44] |
| | | | | RSSCF | √ | √ | | √ | Yan *et al.* 2017 [31] |
| | | | | RS$^2$ACF | √ | √ | | √ | Zhang *et al.* 2019 [33] |
| | | | | MRSCF | √ | √ | | √ | Zhou *et al.* 2020 [131] |
| | | | | SLCF | √ | | | √ | Li *et al.* 2020 [137] |
| | | | | LRCCF | √ | | √ | √ | Shu *et al.* 2015 [30] |
| | | Pairwise constraints based CF | | PCCF | | | | √ | He *et al.* 2014 [25] |
| | | | | SSCF | | | | √ | Lu *et al.* 2016 [26] |
| | | | | CNPCF | √ | | | √ | Lu *et al.* 2016 [27] |
| | | | | RCF | √ | | √ | √ | Yan *et al.* 2017 [28] |
| | Supervised CF | Label information driven | | SGDCF | √ | | | √ | Long *et al.* 2018 [22] |
| **Deep/ multi-layer CF methods** | Traditional feeding-style deep CF model | - | | MCF | | | | | Li *et al.* 2015 [35] |
| | | - | | GMCF | √ | | | | Li *et al.* 2017 [36] |
| | | - | | DGMCF | √ | | | | Zhang *et al.* 2018 [164] |
| | Optimized deep CF model | Single-channel | | DSCF-Net | √ | √ | | | Zhang *et al.* 2019 [37] |
| | | Dual-channel | | DS$^2$CF-Net | √ | √ | | √ | Zhang *et al.* 2020 [38] |

$$(X-E) \leftarrow U_M V$$
$$U_M = U_{M-1} W_M$$
$$\vdots \qquad (64)$$
$$U_2 = U_1 W_2$$
$$U_1 = (X-E) W_1$$

Then, the total reconstruction error between $X-E$ and its reconstruction is $\|(X-E)-(X-E)W_1...W_{M-1}W_M V^T\|_F^2$, where the deep basis is $(X-E)W_0...W_{M-1}W_M$ and the learned deep representation is $V^T$. Inspired by the self-expression of SRMCF [20], DSCF-Net also takes $W_1...W_{M-1}W_M V^T$ as the self-expressive weights and adds the locality constraint $\|V-V(W_1...W_{M-1}W_M V^T)\|_F^2 + \|W_M V^T\|_F^2$ into the CF framework. Finally, the objective function of DSCF-Net is defined as the following joint minimization problem:

$$\min_{W_1,...,W_M,V,E,S'} \|(X-E)-(X-E)W_1...W_{M-1}W_M V^T\|_F^2$$
$$+ \alpha_{DSCF-Net}\left(\|S'-W_1...W_{M-1}W_M V^T\|_F^2 + \|W_M V^T\|_F^2\right), \quad (65)$$
$$+ \beta_{DSCF-Net}\|V^T - V^T S'\|_F^2 + \gamma_{DSCF-Net}\|E\|_{2,1}$$

s.t. $\forall_{m\in\{1,2,...,M\}} W_m \geq 0, V \geq 0$, where the auxiliary variable $S'$ is to facilitate the optimization, $\alpha_{DSCF-Net}$, $\beta_{DSCF-Net}$ and $\gamma_{DSCF-Net}$ are three nonnegative parameters, and the $L_{2,1}$-norm constrained term $\|E\|_{2,1}$ can make the error term $E$ column sparse. It is clear that DSCF-Net can automatically learn the intermediate hidden representations and aim at updating the intermediate basis vectors in each layer. Since the basis vectors can capture the higher-level features and each sample is reconstructed by a linear combination of the bases, optimizing the basis vectors can improve the representation indirectly in each layer.

**DS$^2$CF-Net [38].** Different from DSCF-Net that aims at optimizing the basis vectors to improve the representation indirectly in each layer, DS$^2$CF-Net coupled updates the basis vectors and new representations in each layer. To enhance the representation and clustering abilities, DS$^2$CF-Net designs a hierarchical and coupled factorization framework that has $M$ layers, which is formulated for learning $M$ updated pairs of representation matrices and basis vectors $XW_1...W_M$, and $M$ updated label constraint matrices $A$. The deep factorization process of DS$^2$CF-Net is exhibited in Fig.14, which is performed as follows:

$$X \leftarrow U_M V_M^T$$
$$\begin{array}{ll} U_M = U_{M-1}W_M & V_M = V_{M-1}Z_M \\ \vdots & \vdots \\ U_2 = U_1 W_2 & V_2 = V_1 Z_2 \\ U_1 = XW_1 & V_1 = AZ_1 \end{array} \text{ and } , \quad (66)$$

where $U_m (m=1,2,...,M)$ is the set of basis vectors of the $m$-th layer, $V_m^T (m=1,2,...,M)$ is the low-dimensional representation, $W_m (m=1,2,...,M)$ is the intermediate matrix for updating the basis vectors and $Z_m (m=1,2,...,M)$ is the intermediate auxiliary matrix for updating the low-dimensional representation. Matrix $A$ denotes the label constraint matrix. Finally, the hierarchical model of DS$^2$CF-Net minimizes the objective function:

$$O_{DS^2CF-Net} = \|X - XW_0...W_M (Z_0...Z_M)^T A^T\|_F^2 + \alpha_{DS^2CF-Net} J_1$$
$$+ \beta_{DS^2CF-Net} J_2 + \gamma_{DS^2CF-Net} J_3 \quad , \quad (67)$$
$$s.t. \ \forall_{m\in\{1,2,...,M\}} W_m \geq 0, Z_m \geq 0, S^U \geq 0, S^V \geq 0$$

where $\|X-XW_0...W_M(Z_0...Z_M)^T A^T\|_F^2$ is a deep reconstruction error, $J_1$ is the label propagation function for enriching the prior, $J_2$ is the enriched prior based structure constraint, $J_3$ is the self-weighted dual-graph learning function to obtain the locality preserving representations, $\alpha_{DS^2CF-Net}$, $\beta_{DS^2CF-Net}$ and $\gamma_{DS^2CF-Net}$ are three nonnegative trade-off parameters. Note that both $W_0$ and $Z_0$ are added to facilitate the description, which are fixed to be the identity matrices. The structure of the overall label constraint matrix $A$ is similarly defined as RS$^2$ACF [33].

To enrich the supervised prior information and make full use of both labeled and unlabeled data, DS$^2$CF-Net clearly incorporates the dual constraints, i.e., label constraint and structure constraint. Specifically, the function $J_1$ is defined as

$$J_1 = \|A_L - X_L^T P\|_F^2 + \|P^T X - P^T X(W_0...W_M(Z_0...Z_M)^T A^T)\|_F^2 + \|P\|_{2,1},$$
(68)

where $W_0...W_M(Z_0...Z_M)^T A^T$ can be regarded as the meaningful coefficient matrix self-expressing $X$, based on the self-expression on the coefficient matrix [20]. Since the learned robust label predictor $P$ over $X_L$ can map each $x_i$ into label space by $P^T x_i$, i.e., the supervised information can be enriched by estimating the soft label of each unlabeled sample $x_i \in X_U$ as $x_i^T P$. Then, we obtain $A_U$ using the normalized soft labels that are described as $(A_U)_{ij} = (X_U^T P)_{ij} / \sum_{j=1}^c (X_U^T P)_{ij}$. Then, the enriched prior based structure constraint function $J_2$ can be defined as

$$J_2 = \left[\|\Xi - W_0...W_M(Z_0...Z_M)^T A^T\|_F^2 + \|W_0...W_M(Z_0...Z_M)^T A^T\|_F^2\right]. (69)$$

where the structure-preserving matrix $\Xi$ is defined as follows:

$$\Xi = \begin{bmatrix} \Xi_L & 0 \\ 0 & \Xi_U \end{bmatrix} \in \mathbb{R}^{(l+u)\times(l+u)}, \ \Xi_L = \begin{bmatrix} \Xi_1 & 0 & 0 & 0 \\ 0 & \Xi_2 & 0 & 0 \\ 0 & 0 & ... & 0 \\ 0 & 0 & 0 & \Xi_c \end{bmatrix} \in \mathbb{R}^{l\times l}. \quad (70)$$

where $Q_L$ and $Q_U$ are the structure constraint matrices defined based on $X_L$ and $X_U$, respectively. $Q_L$ is a strict block-diagonal matrix, where each block $Q_i (i=1,2,...,c)$ is a matrix of all ones, defined according to the labeled data. The self-weighted dual-graph learning function $J_3$ is defined as follows:

$$J_3 = \left[\|U_M^T - U_M^T S^U\|_F^2 + \|V_M^T - V_M^T S^V\|_F^2\right], \quad (71)$$

where $U_M = XW_0 \cdots W_M$ and $V_M = A(Z_0...Z_M)$ are auxiliary matrices. Clearly, $J_3$ can effectively retrain the local neighborhood information of the deep basis vectors $XW_0 \cdots W_M$ and representations $(Z_0...Z_M)^T A^T$ in an adaptive manner simultaneously.

**Remarks.** In addition to the factorization model, DS$^2$CF-Net also differs from DSCF-Net in several aspects: 1) DSCF-Net is an unsupervised model, but DS$^2$CF-Net is a semi-supervised method that incorporates the optimized label constraint and label propagation into a unified framework; 2) DS$^2$CF-Net also includes a structure constraint to enforce the deep self-expression weight to be structured and to have a good block-diagonal structure, so that each sample can be reconstructed more accurately by the samples of the same class as much as possible; 3) DSCF-Net preserves the manifold structure information only in data space while DS$^2$CF-Net can not only retain the local manifold information in the data space but also in the resulting feature space by introducing the adaptive dual-graph constraints.

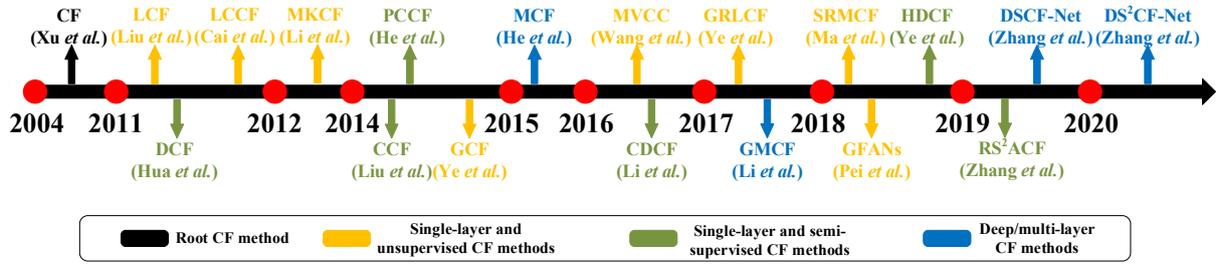

**Fig.15:** Milestone of representative CF methods in each period and their variants.

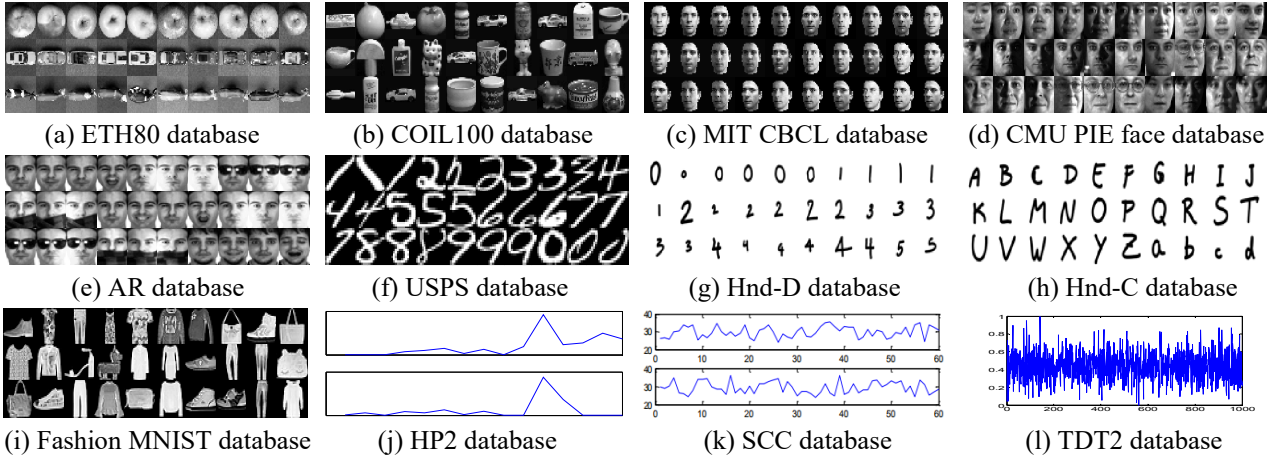

**Fig.16:** Examples of evaluated databases.

**Summarization.** Finally, we summarize all the current CF-based methods in Table II, and also show the milestones of CF-based methods in Fig.15. In Table II, we mainly show the categorization of CF methods and also describe their characteristics. In Fig.15, we mainly show the development trend of CF-based methods, where we clearly illustrate the timeline of representative methods in each period. From the timeline, it is clear that there is an obvious trend in the development of CF methods, i.e., from shallow to deep. From year 2011, many CF variants have begun to be put forward, and in the year 2015 deep CF method was first proposed. From year 2015 till now, deep frameworks have gradually become the mainstream trend.

## V. EXPERIMENTAL RESULTS AND ANALYSIS

We conduct simulations to examine the data representation and clustering performance of some representative CF methods, including the root CF [6], four unsupervised single-layer models (LCCF [7], LCF [16], GCF [8], and RFA-LCF [14]), two semi-supervised single-layer methods (CCF [29] and RS$^2$ACF [33]) and four deep/multi-layer methods (MCF [35], GMCF [36], DSCF-Net [37] and DS$^2$CF-Net [38]). In this study, 12 public real-world databases are involved, including three face databases (i.e., CMU PIE [178], MIT CBCL [172] and AR [171]), two object databases (COIL100 [173] and ETH80 [176]), three handwritten databases (USPS [187], Hand-drawn-digits called Hnd-D [170], and Hand-drawn-characters termed Hnd-C [170]), one fashion product database (Fashion MNIST [188]), one rolling bearing fault dataset (HP2 [189]), one text dataset (TDT2 [193]), and one synthetic control chart time series dataset from UCI (SCC) [190]. Since the dimensionality of the text data in TDT2 [193] is too high, we reduce the dimensionality to 1000 by PCA [1] for efficiency. For the face, object and handwritten digit datasets (Hnd-D and Hnd-C), the images are down-sampled into 32*32 pixels, corresponding to 1024-dimensional vectors. Some samples of evaluated datasets are shown in Fig.16, and the detailed information of the databases is shown in Table III, where we show the total number of samples, dimension and number of classes. We perform all experiments on a PC with Intel Core i5-4590 CPU @ 3.30 GHz 3.30 GHz 8G.

**Table III.** List of used datasets and dataset information.

| Dataset | Data Type | #Points | #Class | #Dim |
|---|---|---|---|---|
| CMU PIE [178] | Face images | 11554 | 68 | 1024 |
| MIT CBCL [172] | | 3240 | 10 | 1024 |
| AR [171] | | 2600 | 100 | 1024 |
| COIL100 [173] | Object images | 7200 | 100 | 1024 |
| ETH80 [176] | | 3280 | 80 | 1024 |
| USPS [187] | Handwritten digits | 3000 | 10 | 256 |
| Hnd-D [170] | | 550 | 10 | 1024 |
| Hnd-C [170] | | 2860 | 52 | 1024 |
| Fashion MNIST [188] | Fashion products | 70000 | 10 | 784 |
| HP2 [189] | Bearing faults | 800 | 10 | 15 |
| TDT2 [193] | Text | 10212 | 96 | 36771 |
| SCC [190] | Time series | 600 | 6 | 60 |

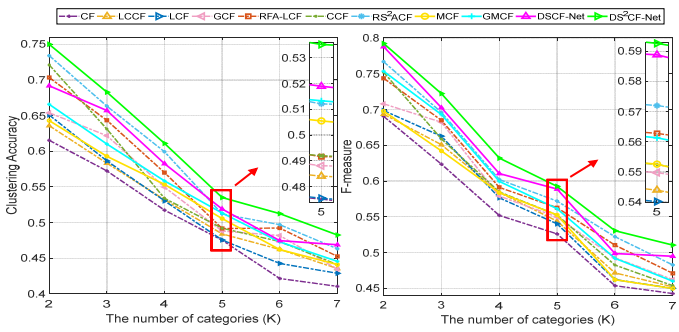
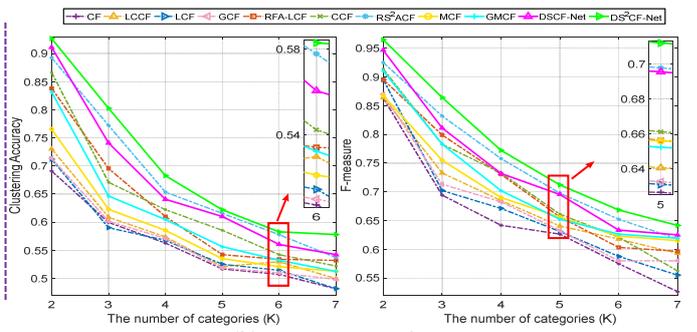

(a) CMU PIE dataset

(b) MIT CBCL dataset

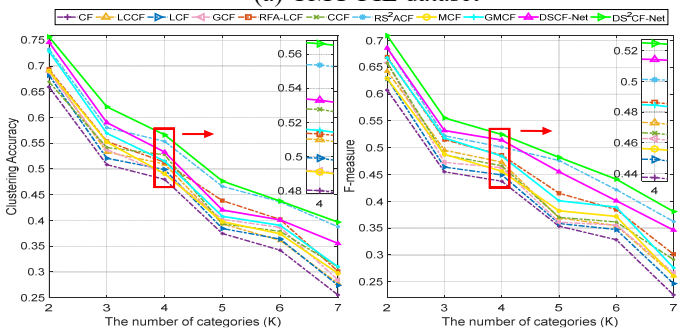
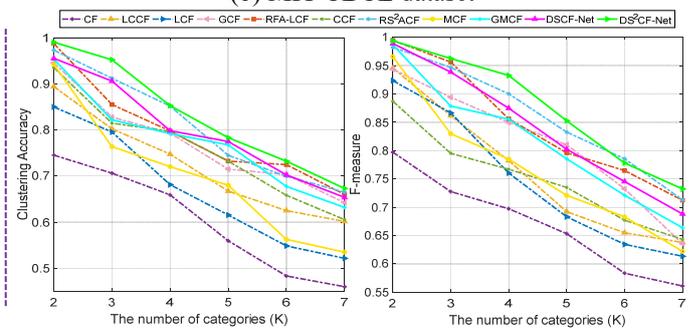

(c) AR dataset

(d) COIL100 dataset

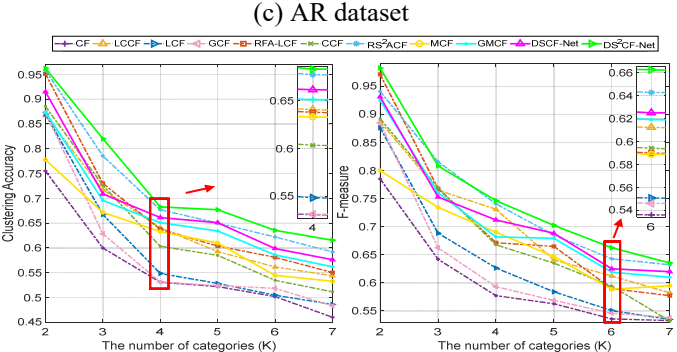
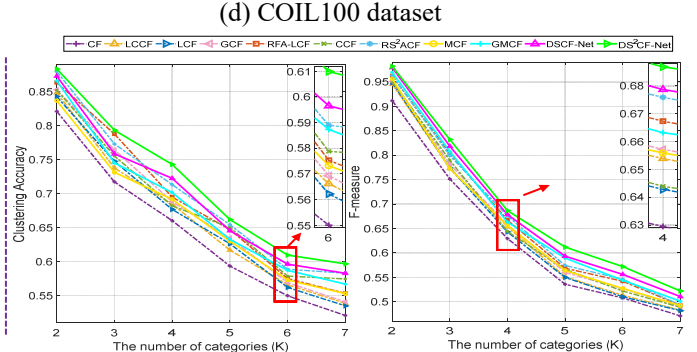

(e) ETH80 dataset

(f) USPS dataset

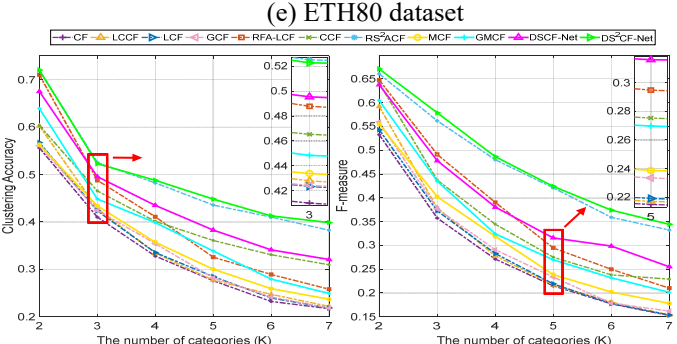
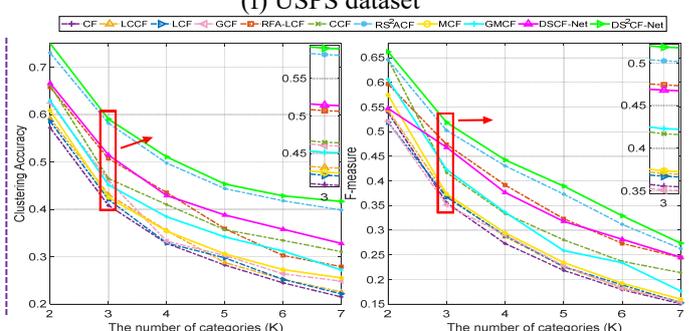

(g) Hnd-D dataset

(h) Hnd-C dataset

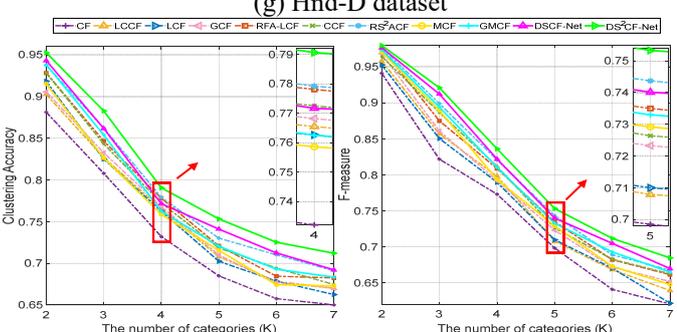
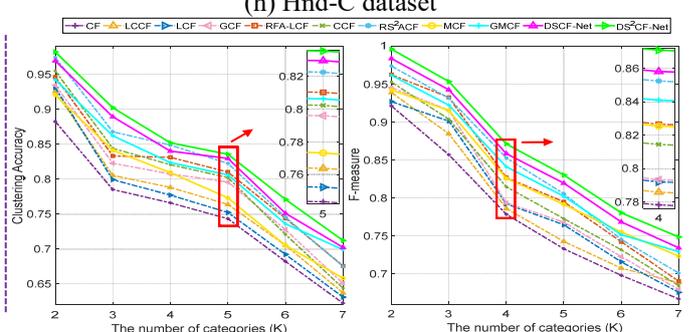

(i) Fashion MNIST dataset

(j) HP2 dataset

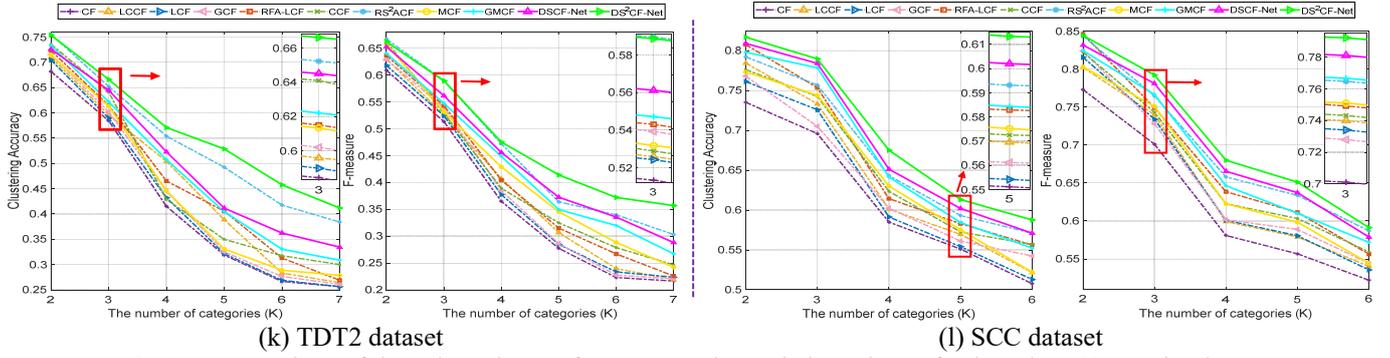

(k) TDT2 dataset  (l) SCC dataset

**Fig.17:** Comparison of data clustering performance under varied numbers of K based on 12 popular datasets.

## A. Clustering Evaluation Process and Metrices

***Clustering Evaluation Process.*** For the quantitative clustering evaluations, we perform K-means with cosine distance on the new representations obtained by each method. Following the common procedures in [7-8][16][14][29][33], for each number K of clusters, we randomly choose K categories from each set and use the data of K categories to form the matrix *X*. To avoid randomness, we average the highest five times numerical results over 20 random selections of K categories for each evaluated algorithm. Note that the rank *r* of the matrix is set to K+1 according to [174] for all the evaluated CF models for fair comparison. For semi-supervised methods CCF [29] and RS$^2$ACF [33], the proportion of labeled samples is set to 25%.

***Evaluation Metrics.*** We employ two widely-used quantitative evaluation metrics, i.e., Clustering Accuracy (AC) and F-measure [179-180]. AC is the percentage of the cluster labels to the

**Table IV.** Averaged clustering accuracies (AC) of each algorithm based on the evaluated real databases.

| Methods | | CMU PIE | | MIT CBCL | | AR | | COIL100 | |
|---|---|---|---|---|---|---|---|---|---|
| | | Mean±std | Best | Mean±std | Best | Mean±std | Best | Mean±std | Best |
| Single-layer methods | CF | 0.5021±0.0820 | 0.6153 | 0.5598±0.0769 | 0.6909 | 0.4366±0.1431 | 0.6595 | 0.6023±0.1189 | 0.7453 |
| | LCCF | 0.5222±0.0765 | 0.6357 | 0.5772±0.0847 | 0.7303 | 0.4611±0.1477 | 0.6919 | 0.7229±0.1127 | 0.8938 |
| | LCF | 0.5191±0.0871 | 0.6508 | 0.5652±0.0816 | 0.7120 | 0.4540±0.1442 | 0.6821 | 0.6687±0.1327 | 0.8498 |
| | GCF | 0.5383±0.0859 | 0.6538 | 0.5688±0.0812 | 0.7136 | 0.4650±0.1404 | 0.6882 | 0.7711±0.1082 | 0.9455 |
| | RFA-LCF | 0.5588±0.0984 | 0.7033 | 0.6253±0.1219 | 0.8380 | 0.4837±0.1355 | 0.6937 | 0.7929±0.1163 | 0.9872 |
| | CCF | 0.5487±0.1070 | 0.7210 | 0.6349±0.1255 | 0.8662 | 0.4701±0.1327 | 0.6692 | 0.7571±0.1183 | 0.9350 |
| | RS$^2$ACF | **0.5781±0.1060** | **0.7340** | **0.6748±0.1336** | **0.8925** | **0.5261±0.1242** | **0.7325** | **0.8076±0.1234** | **0.9728** |
| Multi-layer methods | MCF | 0.5330±0.0779 | 0.6430 | 0.5905±0.0956 | 0.7655 | 0.4672±0.1415 | 0.6907 | 0.7006±0.1480 | 0.9420 |
| | GMCF | 0.5443±0.0838 | 0.6658 | 0.6139±0.1173 | 0.8315 | 0.4870±0.1501 | 0.7285 | 0.7740±0.1134 | 0.9532 |
| | DSCF-Net | 0.5656±0.0942 | 0.6918 | 0.6676±0.1385 | 0.9111 | 0.5077±0.1460 | 0.7462 | 0.7984±0.1156 | 0.9550 |
| | DS$^2$CF-Net | **0.5957±0.1049** | **0.7500** | **0.6991±0.1392** | **0.9267** | **0.5424±0.1335** | **0.7566** | **0.8305±0.1244** | **0.9899** |
| | | ETH80 | | USPS | | Hnd-D | | Hnd-C | |
| | | Mean±std | Best | Mean±std | Best | Mean±std | Best | Mean±std | Best |
| Single-layer methods | CF | 0.5616±0.1052 | 0.7552 | 0.6439±0.1127 | 0.8210 | 0.3366±0.1284 | 0.5561 | 0.3417±0.1316 | 0.5720 |
| | LCCF | 0.6538±0.1217 | 0.8671 | 0.6655±0.1160 | 0.8485 | 0.3518±0.1429 | 0.6022 | 0.3574±0.1371 | 0.5924 |
| | LCF | 0.6007±0.1465 | 0.8698 | 0.6652±0.1162 | 0.8428 | 0.3448±0.1314 | 0.5672 | 0.3512±0.1341 | 0.5859 |
| | GCF | 0.5930±0.1455 | 0.8730 | 0.6752±0.1188 | 0.8529 | 0.3470±0.1293 | 0.5609 | 0.3722±0.1449 | 0.6244 |
| | RFA-LCF | 0.6756±0.1485 | 0.9513 | 0.6871±0.1210 | 0.8633 | 0.4137±0.1678 | 0.7102 | 0.4235±0.1427 | 0.6580 |
| | CCF | 0.6410±0.1412 | 0.8855 | 0.6823±0.1072 | 0.8522 | 0.4119±0.1088 | 0.6033 | 0.4231±0.1296 | 0.6621 |
| | RS$^2$ACF | **0.7141±0.1369** | **0.9584** | **0.6987±0.1147** | **0.8796** | **0.4924±0.1221** | **0.7189** | **0.5116±0.1257** | **0.7300** |
| Multi-layer methods | MCF | 0.6282±0.0902 | 0.7778 | 0.6694±0.1065 | 0.8367 | 0.3585±0.1228 | 0.5630 | 0.3702±0.1318 | 0.6084 |
| | GMCF | 0.6671±0.1121 | 0.8744 | 0.6833±0.1121 | 0.8660 | 0.3922±0.1416 | 0.6391 | 0.3985±0.1289 | 0.6295 |
| | DSCF-Net | 0.6852±0.1220 | 0.9148 | 0.6966±0.1105 | 0.8729 | 0.4417±0.1309 | 0.6752 | 0.4473±0.1253 | 0.6662 |
| | DS$^2$CF-Net | **0.7322±0.1335** | **0.9623** | **0.7147±0.1123** | **0.8833** | **0.4984±0.1182** | **0.7203** | **0.5249±0.1277** | **0.7503** |
| | | Fashion MNIST | | HP2 | | TDT2 | | SCC | |
| | | Mean±std | Best | Mean±std | Best | Mean±std | Best | Mean±std | Best |
| Single-layer methods | CF | 0.7357±0.0920 | 0.8812 | 0.7470±0.0893 | 0.8825 | 0.4205±0.1768 | 0.6820 | 0.6149±0.0968 | 0.7352 |
| | LCCF | 0.7586±0.0919 | 0.9026 | 0.7710±0.0975 | 0.9255 | 0.4577±0.1775 | 0.7102 | 0.6420±0.1118 | 0.7844 |
| | LCF | 0.7583±0.0987 | 0.9183 | 0.7639±0.1022 | 0.9310 | 0.4288±0.1837 | 0.7045 | 0.6291±0.1088 | 0.7612 |
| | GCF | 0.7605±0.0938 | 0.9055 | 0.7899±0.0955 | 0.9346 | 0.4375±0.1866 | 0.7153 | 0.6358±0.0970 | 0.7679 |
| | RFA-LCF | 0.7735±0.0982 | 0.9282 | 0.8068±0.0915 | 0.9463 | 0.4652±0.1757 | 0.7233 | 0.6629±0.1110 | **0.8075** |
| | CCF | 0.7719±0.0979 | 0.9284 | 0.7978±0.1067 | 0.9550 | 0.4621±0.1832 | 0.7358 | 0.6543±0.0997 | 0.7762 |
| | RS$^2$ACF | **0.7851±0.0966** | **0.9375** | **0.8223±0.1034** | **0.9742** | **0.5427±0.1420** | **0.7557** | **0.6713±0.0986** | 0.7922 |
| Multi-layer methods | MCF | 0.7605±0.0953 | 0.9152 | 0.7846±0.0945 | 0.9205 | 0.4454±0.1823 | 0.7145 | 0.6489±0.1076 | 0.7731 |
| | GMCF | 0.7753±0.1011 | 0.9381 | 0.8116±0.0873 | 0.9422 | 0.4845±0.1689 | 0.7338 | 0.6709±0.1120 | 0.7980 |
| | DSCF-Net | 0.7870±0.0968 | 0.9431 | 0.8303±0.0954 | 0.9693 | 0.5004±0.1593 | 0.7260 | 0.6834±0.1075 | 0.8091 |
| | DS$^2$CF-Net | **0.8028±0.0955** | **0.9525** | **0.8425±0.0951** | **0.9820** | **0.5649±0.1283** | 0.7540 | **0.6965±0.1029** | **0.8168** |

**Table V.** Averaged F-scores of each algorithm based on the evaluated real databases.

| Methods | | CMU PIE | | MIT CBCL | | AR | | COIL100 | |
|---|---|---|---|---|---|---|---|---|---|
| | | Mean±std | Best | Mean±std | Best | Mean±std | Best | Mean±std | Best |
| Single-layer methods | CF | 0.5482±0.0964 | 0.6905 | 0.6545±0.1174 | 0.8633 | 0.4010±0.1307 | 0.6075 | 0.6700±0.0892 | 0.7971 |
| | LCCF | 0.5661±0.0960 | 0.6935 | 0.6889±0.0993 | 0.8643 | 0.4326±0.1336 | 0.6429 | 0.7624±0.1231 | 0.9455 |
| | LCF | 0.5650±0.1019 | 0.6986 | 0.6737±0.1210 | 0.8951 | 0.4156±0.1309 | 0.6292 | 0.7469±0.1265 | 0.9233 |
| | GCF | 0.5790±0.0992 | 0.7080 | 0.6764±0.1091 | 0.8702 | 0.4293±0.1357 | 0.6582 | 0.8109±0.1122 | 0.9436 |
| | RFA-LCF | 0.5940±0.1037 | 0.7437 | 0.7135±0.1181 | 0.8953 | 0.4618±0.1269 | 0.6694 | 0.8466±0.1107 | **0.9944** |
| | CCF | 0.5808±0.1114 | 0.7522 | 0.7123±0.1260 | 0.9138 | 0.4388±0.1293 | 0.6579 | 0.7508±0.0872 | 0.8870 |
| | RS²ACF | **0.6070±0.1071** | **0.7669** | **0.7471±0.1158** | **0.9250** | **0.4949±0.1104** | **0.6867** | **0.8599±0.1019** | 0.9820 |
| Multi-layer methods | MCF | 0.5647±0.0979 | 0.6977 | 0.7011±0.0963 | 0.8677 | 0.4311±0.1250 | 0.6291 | 0.7677±0.1215 | 0.9655 |
| | GMCF | 0.5933±0.1137 | 0.7535 | 0.7155±0.1132 | 0.9105 | 0.4560±0.1338 | 0.6679 | 0.8151±0.1163 | 0.9866 |
| | DSCF-Net | 0.6139±0.1150 | 0.7878 | 0.7406±0.1220 | 0.9466 | 0.4890±0.1188 | 0.6860 | 0.8396±0.1153 | 0.9887 |
| | DS²CF-Net | **0.6300±0.1099** | **0.7922** | **0.7705±0.1241** | **0.9648** | **0.5154±0.1131** | **0.7091** | **0.8752±0.1051** | **0.9935** |

| Methods | | ETH80 | | USPS | | Hnd-D | | Hnd-C | |
|---|---|---|---|---|---|---|---|---|---|
| | | Mean±std | Best | Mean±std | Best | Mean±std | Best | Mean±std | Best |
| Single-layer methods | CF | 0.6060±0.0962 | 0.7850 | 0.6345±0.1690 | 0.9115 | 0.2844±0.1420 | 0.5328 | 0.2861±0.1439 | 0.5393 |
| | LCCF | 0.7031±0.1141 | 0.8869 | 0.6533±0.1783 | 0.9456 | 0.3001±0.1640 | 0.5921 | 0.2942±0.1445 | 0.5436 |
| | LCF | 0.6435±0.1268 | 0.8761 | 0.6513±0.1803 | 0.9488 | 0.2915±0.1463 | 0.5430 | 0.2905±0.1354 | 0.5191 |
| | GCF | 0.6322±0.1313 | 0.8842 | 0.6613±0.1766 | 0.9510 | 0.3002±0.1493 | 0.5593 | 0.2876±0.1349 | 0.5206 |
| | RFA-LCF | 0.7070±0.1463 | **0.9710** | 0.6728±0.1797 | 0.9626 | 0.3806±0.1656 | 0.6476 | 0.3839±0.1334 | 0.5968 |
| | CCF | 0.6817±0.1298 | 0.8920 | 0.6602±0.1783 | 0.9511 | 0.3609±0.1577 | 0.6410 | 0.3547±0.1605 | **0.6467** |
| | RS²ACF | **0.7427±0.1183** | 0.9405 | **0.6753±0.1777** | **0.9628** | **0.4691±0.1253** | **0.6605** | **0.4211±0.1390** | 0.6460 |
| Multi-layer methods | MCF | 0.6758±0.0824 | 0.7998 | 0.6614±0.1762 | 0.9561 | 0.3159±0.1431 | 0.5548 | 0.3048±0.1524 | 0.5744 |
| | GMCF | 0.7126±0.1172 | 0.9250 | 0.6791±0.1784 | 0.9696 | 0.3445±0.1517 | 0.6046 | 0.3391±0.1564 | 0.6060 |
| | DSCF-Net | 0.7215±0.1146 | 0.9310 | 0.6894±0.1789 | 0.9795 | 0.3943±0.1422 | 0.6377 | 0.3727±0.1160 | 0.5473 |
| | DS²CF-Net | **0.7561±0.1266** | **0.9821** | **0.7012±0.1749** | **0.9820** | **0.4797±0.1258** | **0.6710** | **0.4539±0.1400** | **0.6622** |

| Methods | | Fashion MNIST | | HP2 | | TDT2 | | SCC | |
|---|---|---|---|---|---|---|---|---|---|
| | | Mean±std | Best | Mean±std | Best | Mean±std | Best | Mean±std | Best |
| Single-layer methods | CF | 0.7495±0.1211 | 0.9411 | 0.7756±0.0974 | 0.9210 | 0.3672±0.1621 | 0.6088 | 0.6268±0.1059 | 0.7730 |
| | LCCF | 0.7716±0.1224 | 0.9582 | 0.7912±0.1004 | 0.9386 | 0.3879±0.1643 | 0.6295 | 0.6522±0.1128 | 0.8023 |
| | LCF | 0.7659±0.1228 | 0.9524 | 0.7959±0.1001 | 0.9268 | 0.3767±0.1627 | 0.6177 | 0.6534±0.1169 | 0.8155 |
| | GCF | 0.7782±0.1211 | 0.9682 | 0.8040±0.1058 | 0.9453 | 0.3812±0.1703 | 0.6291 | 0.6574±0.1154 | 0.8239 |
| | RFA-LCF | 0.7897±0.1209 | 0.9755 | 0.8247±0.1061 | 0.9622 | 0.3985±0.1622 | 0.6363 | 0.6803±0.1163 | **0.8462** |
| | CCF | 0.7892±0.1199 | 0.9644 | 0.8098±0.1024 | 0.9525 | 0.4050±0.1608 | 0.6622 | 0.6701±0.1089 | 0.8222 |
| | RS²ACF | **0.7996±0.1220** | **0.9768** | **0.8348±0.1055** | **0.9735** | **0.4558±0.1470** | **0.6669** | **0.6974±0.1045** | 0.8437 |
| Multi-layer methods | MCF | 0.7840±0.1264 | 0.9706 | 0.8254±0.0874 | 0.9422 | 0.4156±0.1572 | 0.6577 | 0.6639±0.1085 | 0.8028 |
| | GMCF | 0.7950±0.1210 | 0.9738 | 0.8345±0.0926 | 0.9615 | 0.4286±0.1439 | 0.6402 | 0.6839±0.1072 | 0.8243 |
| | DSCF-Net | 0.8043±0.1214 | 0.9766 | 0.8509±0.0974 | 0.9835 | 0.4442±0.1404 | 0.6524 | 0.6988±0.1045 | 0.8316 |
| | DS²CF-Net | **0.8145±0.1185** | **0.9798** | **0.8631±0.0969** | **0.9959** | **0.4782±0.1234** | **0.6622** | **0.7117±0.1040** | **0.8443** |

true labels provided by the original data corpus:

$$AC = \sum_{i=1}^{N} \delta(r_i, map(p_i)) / N, \quad (72)$$

where $N$ is the number of samples, and the function $map(p_i)$ denotes the permutation mapping function that maps the cluster label $p_i$ obtained by the clustering method to the true label $r_i$ provided by the original data corpus.

The F-measure metric for clustering is defined as follows:

$$F_\mu = \frac{(\mu^2 + 1) PRECISION \times RECALL}{\mu^2 PRECISION + RECALL}, \quad (73)$$

where we set the parameter $\mu = 1$ in the following simulations.

### B. Quantitative Clustering Results and Analysis

***Clustering with different values of K.*** We first discuss the clustering performance of each evaluated CF model under varied K numbers. In this study, the number of layers for all multi-layer methods MCF, GMCF, DSCF-Net and DS²CF-Net is set to 3. Since the SCC database only contains 6 categories, we vary the value of K from 2 to 6 over SCC, and for other evaluated databases we vary the value of K from 2 to 7 with step 1. The clustering curves over each database are shown in Fig.17. Note that the averaged values of AC and F-measure according to the curves in Fig.17 are summarized in Tables IV-V, respectively. From the results, we see that: (1) the values of AC and F-measure of each algorithm go down as the number of categories increases, which is easy to understand, because clustering data of less categories is easier than clustering more; (2) in most cases, semi-supervised single-layer methods can deliver better clustering performance than those unsupervised single-layer methods, which is also easy to understand, since semi-supervised also use partial labeled data. RS²ACF obtains the best records among all the semi-supervised methods; (3) MCF and GMCF can respectively perform better than CF and LCCF in investigated cases, which indicates that deep models can indeed improve the representation and clustering power than traditional shallow methods. DS²CF-Net obtains the highest AC and F-measure in most cases, compared with other competitors, which can be benefited by its coupled deep factorization structure and semi-supervised

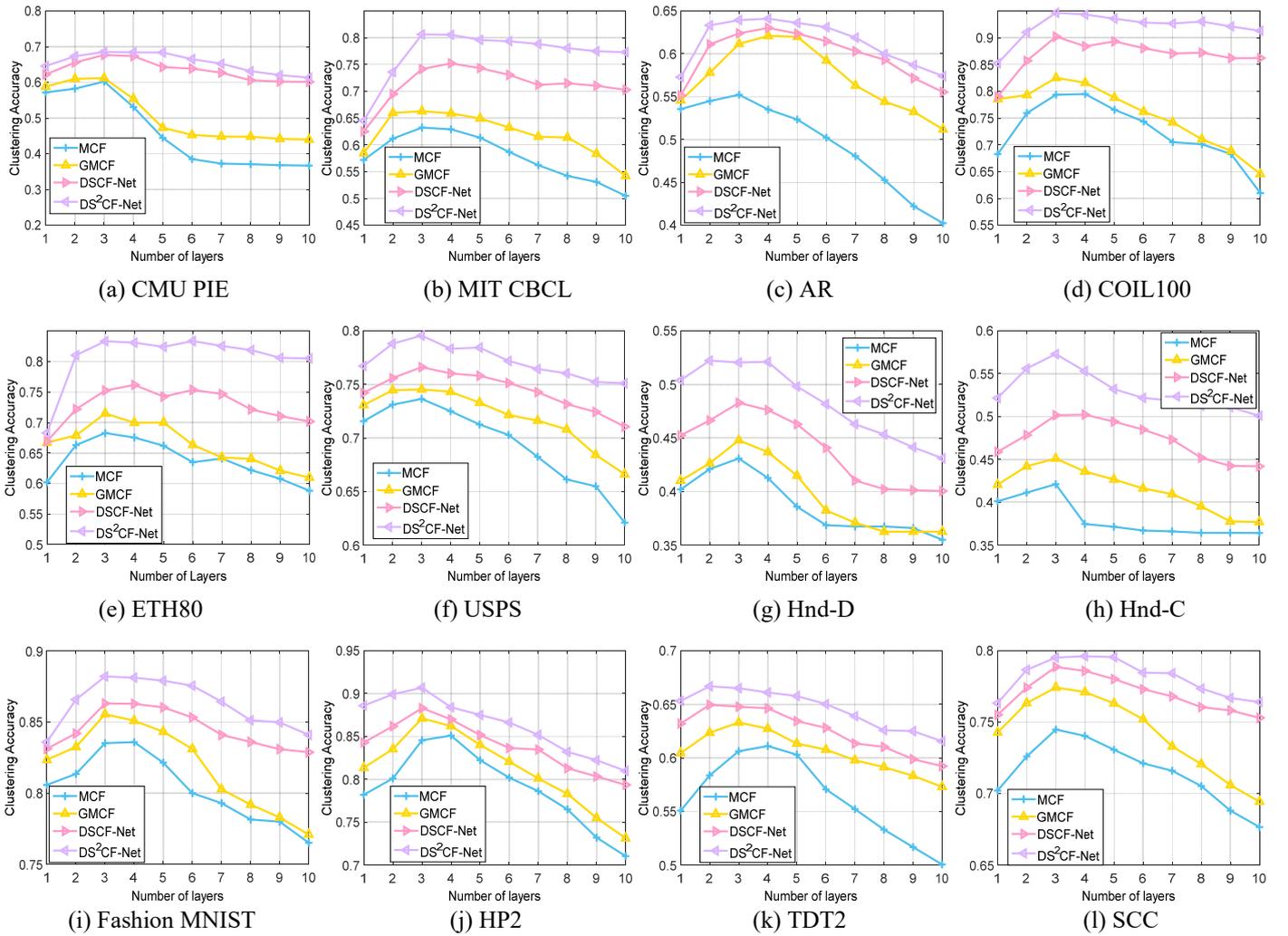

**Fig.18:** Clustering performance in terms of AC under varied number of layers, with K=3, for multi-layer CF methods.

**Table VI.** Averaged clustering accuracies (AC) of each algorithm based on the evaluated real databases (K=3).

| Methods | CMU PIE | | MIT CBCL | | AR | | COIL100 | |
|---|---|---|---|---|---|---|---|---|
| | Mean±std | Best | Mean±std | Best | Mean±std | Best | Mean±std | Best |
| MCF | 0.4596±0.1006 | 0.6022 | 0.5785±0.0436 | 0.6323 | 0.4952±0.0535 | 0.5523 | 0.7239±0.0582 | 0.7950 |
| GMCF | 0.5068±0.0746 | 0.6120 | 0.6202±0.0403 | 0.6628 | 0.5719±0.0386 | 0.6208 | 0.7557±0.0583 | 0.8250 |
| DSCF-Net | 0.6346±0.0279 | 0.6767 | 0.7128±0.0361 | 0.7522 | 0.5978±0.0289 | 0.6302 | 0.8673±0.0306 | 0.9020 |
| **DS²CF-Net** | **0.6553±0.0271** | **0.6850** | **0.7698±0.0483** | **0.8062** | **0.6133±0.0274** | **0.6410** | **0.9206±0.0268** | **0.9466** |
| | ETH80 | | USPS | | Hnd-D | | Hnd-C | |
| | Mean±std | Best | Mean±std | Best | Mean±std | Best | Mean±std | Best |
| MCF | 0.6380±0.0326 | 0.6828 | 0.6943±0.0381 | 0.7364 | 0.3877±0.0269 | 0.4308 | 0.3806±0.0218 | 0.4210 |
| GMCF | 0.6639±0.0353 | 0.7150 | 0.7193±0.0266 | 0.7453 | 0.3977±0.0333 | 0.4478 | 0.4152±0.0255 | 0.4510 |
| DSCF-Net | 0.7283±0.0288 | 0.7618 | 0.7443±0.0177 | 0.7661 | 0.4396±0.0331 | 0.4828 | 0.4729±0.0231 | 0.5022 |
| **DS²CF-Net** | **0.8072±0.0449** | **0.8336** | **0.7718±0.0154** | **0.7955** | **0.4836±0.0346** | **0.5222** | **0.5300±0.0232** | **0.5728** |
| | Fashion MNIST | | HP2 | | TDT2 | | SCC | |
| | Mean±std | Best | Mean±std | Best | Mean±std | Best | Mean±std | Best |
| MCF | 0.8012±0.0229 | 0.8359 | 0.7897±0.0453 | 0.8510 | 0.5628±0.0387 | 0.6111 | 0.7151±0.0221 | 0.7448 |
| GMCF | 0.8196±0.0307 | 0.8556 | 0.8113±0.0448 | 0.8710 | 0.6054±0.0195 | 0.6330 | 0.7420±0.0279 | 0.7743 |
| DSCF-Net | 0.8450±0.0137 | 0.8631 | 0.8390±0.0291 | 0.8829 | 0.6252±0.0206 | 0.6495 | 0.7696±0.0128 | 0.7886 |
| **DS²CF-Net** | **0.8626±0.0172** | **0.8821** | **0.8633±0.0332** | **0.9068** | **0.6459±0.0184** | **0.6668** | **0.7810±0.0131** | **0.7961** |

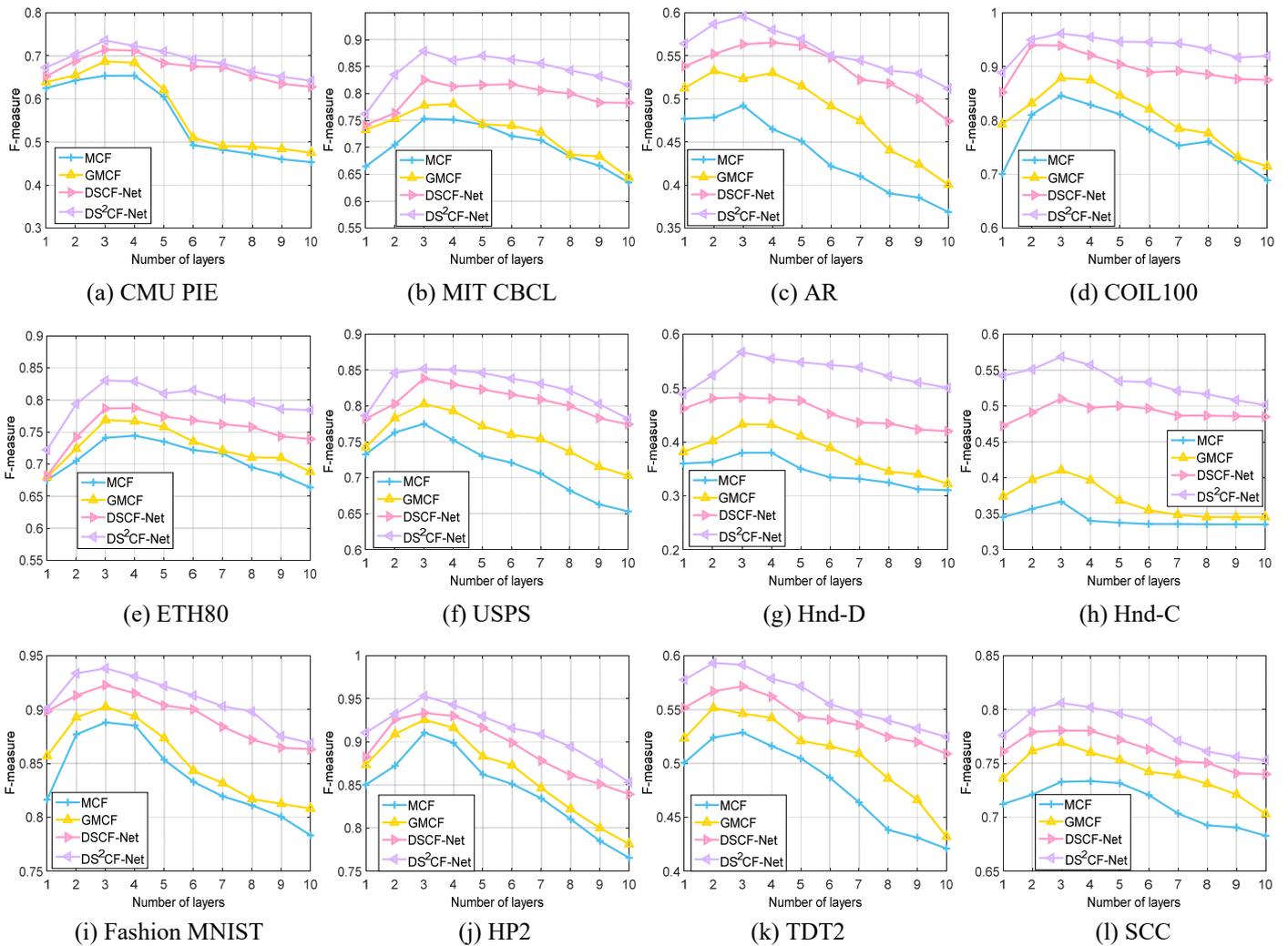

**Fig.19:** Clustering performance in terms of F-measure under varied number of layers, with K=3, for multi-layer CF methods.

**Table VII.** Averaged F-scores of each algorithm based on the evaluated real databases (K=3).

| Methods | CMU PIE | | MIT CBCL | | AR | | COIL100 | |
|---|---|---|---|---|---|---|---|---|
| | Mean±std | Best | Mean±std | Best | Mean±std | Best | Mean±std | Best |
| MCF | 0.5544±0.0882 | 0.6540 | 0.7035±0.0404 | 0.7530 | 0.4341±0.0443 | 0.4922 | 0.7710±0.0543 | 0.8462 |
| GMCF | 0.5735±0.0906 | 0.6870 | 0.7270±0.0435 | 0.7805 | 0.4845±0.0476 | 0.5326 | 0.8055±0.0556 | 0.8789 |
| DSCF-Net | 0.6715±0.0296 | 0.7143 | 0.7949±0.0267 | 0.8252 | 0.5344±0.0303 | 0.5655 | 0.8977±0.0285 | 0.9400 |
| **DS$^2$CF-Net** | **0.6879±0.0308** | **0.7360** | **0.8418±0.0340** | **0.8789** | **0.5566±0.0273** | **0.5962** | **0.9360±0.0218** | **0.9615** |
| | ETH80 | | USPS | | Hnd-D | | Hnd-C | |
| | Mean±std | Best | Mean±std | Best | Mean±std | Best | Mean±std | Best |
| MCF | 0.7081±0.0284 | 0.7444 | 0.7180±0.0415 | 0.7752 | 0.3445±0.0258 | 0.3802 | 0.3424±0.0110 | 0.3668 |
| GMCF | 0.7262±0.0310 | 0.7685 | 0.7563±0.0327 | 0.8030 | 0.3817±0.0386 | 0.4328 | 0.3687±0.0249 | 0.4105 |
| DSCF-Net | 0.7544±0.0309 | 0.7877 | 0.8059±0.0215 | 0.8384 | 0.4547±0.0250 | 0.4825 | 0.4912±0.0104 | 0.5102 |
| **DS$^2$CF-Net** | **0.7969±0.0310** | **0.8305** | **0.8257±0.0264** | **0.8520** | **0.5296±0.0249** | **0.5668** | **0.5334±0.0218** | **0.5687** |
| | Fashion MNIST | | HP2 | | TDT2 | | SCC | |
| | Mean±std | Best | Mean±std | Best | Mean±std | Best | Mean±std | Best |
| MCF | 0.8367±0.0372 | 0.8880 | 0.8440±0.0466 | 0.9105 | 0.4814±0.0402 | 0.5286 | 0.7121±0.0189 | 0.7335 |
| GMCF | 0.8531±0.0360 | 0.9025 | 0.8629±0.0495 | 0.9255 | 0.5093±0.0380 | 0.5515 | 0.7416±0.0203 | 0.7692 |
| DSCF-Net | 0.8936±0.0216 | 0.9226 | 0.8917±0.0344 | 0.9335 | 0.5424±0.0209 | 0.5716 | 0.7619±0.0158 | 0.7805 |
| **DS$^2$CF-Net** | **0.9083±0.0238** | **0.9382** | **0.9115±0.0308** | **0.9531** | **0.5611±0.0247** | **0.5932** | **0.7809±0.0200** | **0.8061** |

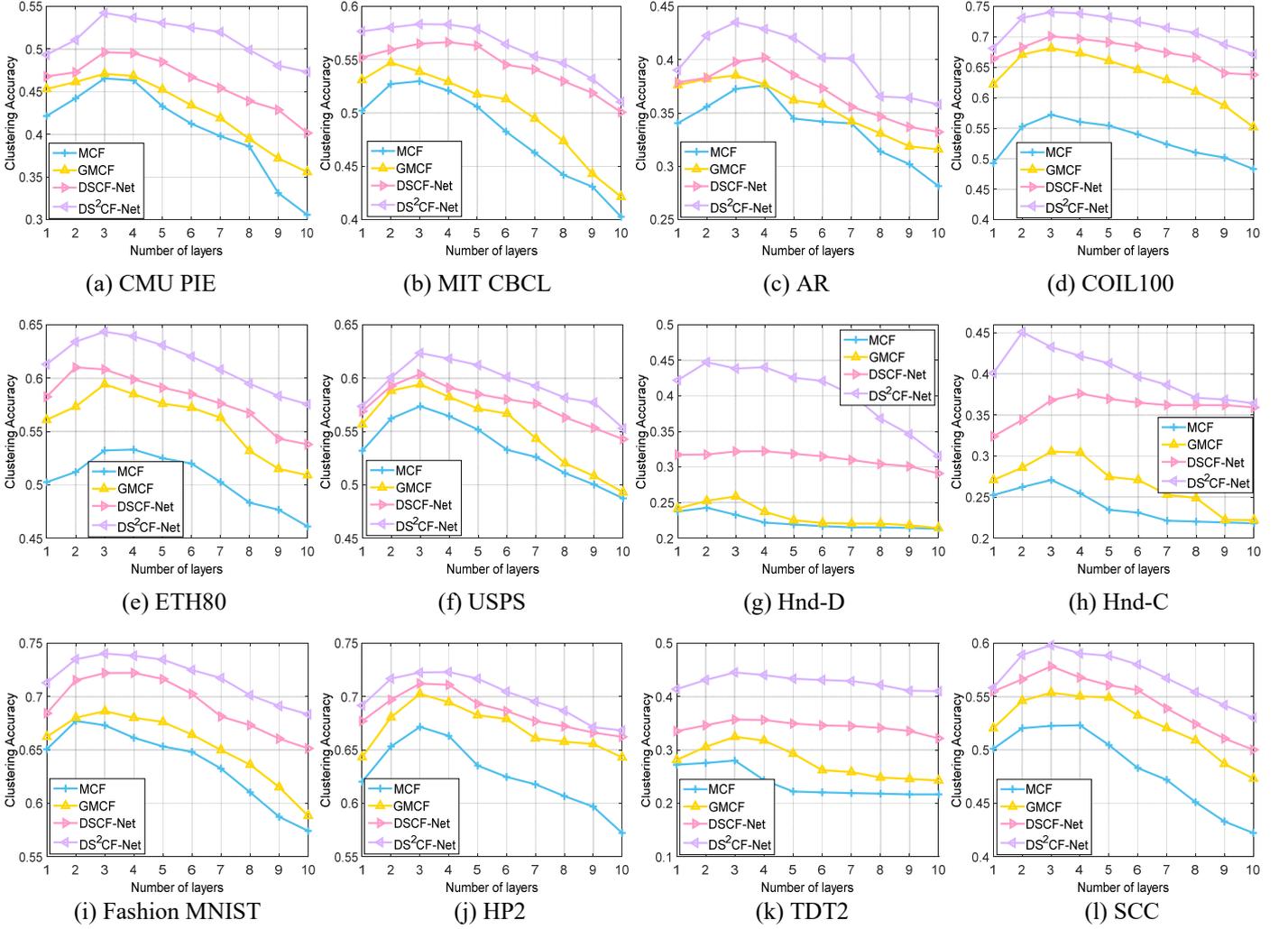

**Fig.20:** Clustering performance in terms of AC under varied number of layers, with K=6, for multi-layer CF methods.

**Table VIII.** Averaged clustering accuracies (AC) of each algorithm based on the evaluated real databases (K=6).

| Methods | CMU PIE | | MIT CBCL | | AR | | COIL100 | |
|---|---|---|---|---|---|---|---|---|
| | Mean±std | Best | Mean±std | Best | Mean±std | Best | Mean±std | Best |
| **MCF** | 0.4059±0.0979 | 0.4655 | 0.4808±0.0443 | 0.5299 | 0.3369±0.0300 | 0.3760 | 0.5292±0.0311 | 0.5722 |
| **GMCF** | 0.4282±0.1137 | 0.4710 | 0.5011±0.0423 | 0.5473 | 0.3548±0.0263 | 0.3853 | 0.6332±0.0417 | 0.6813 |
| **DSCF-Net** | 0.4608±0.1150 | 0.4962 | 0.5442±0.0219 | 0.5664 | 0.3692±0.0249 | 0.4020 | 0.6738±0.0218 | 0.7010 |
| **DS²CF-Net** | **0.5111±0.1099** | **0.5422** | **0.5609±0.0249** | **0.5833** | **0.3988±0.0285** | **0.4350** | **0.7127±0.0251** | **0.7405** |
| | ETH80 | | USPS | | Hnd-D | | Hnd-C | |
| | Mean±std | Best | Mean±std | Best | Mean±std | Best | Mean±std | Best |
| **MCF** | 0.5048±0.0245 | 0.5330 | 0.5341±0.0289 | 0.5736 | 0.2230±0.0108 | 0.2428 | 0.2385±0.0199 | 0.2710 |
| **GMCF** | 0.5580±0.0294 | 0.5943 | 0.5524±0.0351 | 0.5941 | 0.2310±0.0156 | 0.2588 | 0.2658±0.0296 | 0.3055 |
| **DSCF-Net** | 0.5800±0.0247 | 0.6100 | 0.5757±0.0189 | 0.6038 | 0.3119±0.0103 | 0.3224 | 0.3593±0.0149 | 0.3762 |
| **DS²CF-Net** | **0.6142±0.0237** | **0.6435** | **0.5933±0.0222** | **0.6234** | **0.4026±0.0446** | **0.4474** | **0.4007±0.0291** | **0.4511** |
| | Fashion MNIST | | HP2 | | TDT2 | | SCC | |
| | Mean±std | Best | Mean±std | Best | Mean±std | Best | Mean±std | Best |
| **MCF** | 0.6367±0.0204 | 0.6771 | 0.6261±0.0307 | 0.6716 | 0.2383±0.0271 | 0.2800 | 0.4832±0.0376 | 0.5230 |
| **GMCF** | 0.6538±0.0263 | 0.6862 | 0.6699±0.0208 | 0.7025 | 0.2778±0.0309 | 0.3246 | 0.5241±0.0278 | 0.5535 |
| **DSCF-Net** | 0.6929±0.0321 | 0.7221 | 0.6854±0.0176 | 0.7122 | 0.3432±0.0106 | 0.3568 | 0.5457±0.0263 | 0.5784 |
| **DS²CF-Net** | **0.7179±0.0354** | **0.7402** | **0.6997±0.0204** | **0.7230** | **0.4266±0.0122** | **0.4452** | **0.5697±0.0230** | **0.5982** |

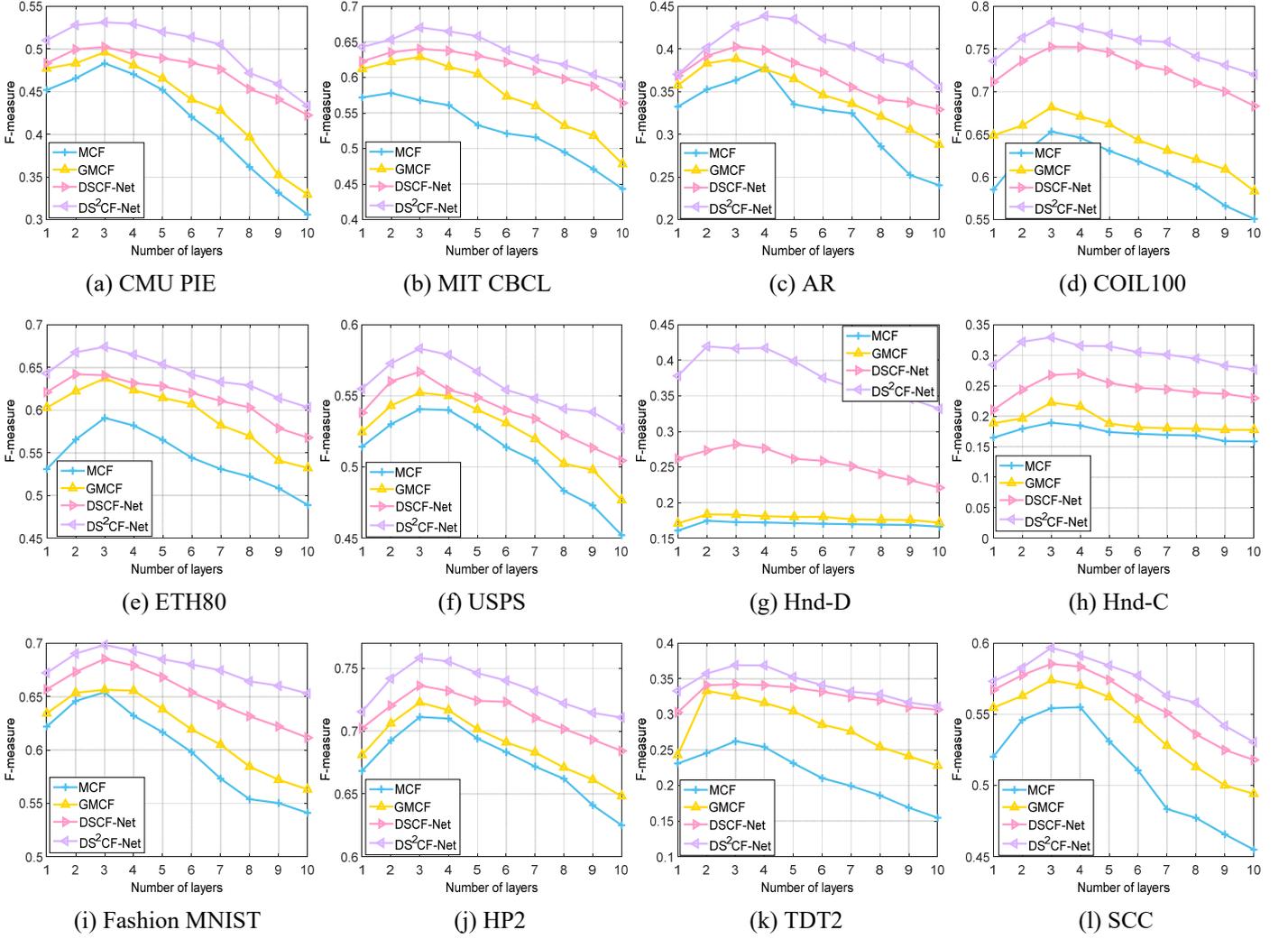

**Fig.21:** Clustering performance in terms of F-measure under varied number of layers, with K=6, for multi-layer CF methods.

**Table IX.** Averaged F-scores of each algorithm based on the evaluated real databases (K=6).

| Methods | CMU PIE | | MIT CBCL | | AR | | COIL100 | |
|---|---|---|---|---|---|---|---|---|
| | Mean±std | Best | Mean±std | Best | Mean±std | Best | Mean±std | Best |
| MCF | 0.4139±0.0624 | 0.4833 | 0.5259±0.0457 | 0.5782 | 0.3195±0.0458 | 0.3782 | 0.6071±0.0339 | 0.6533 |
| GMCF | 0.4351±0.0581 | 0.4962 | 0.5745±0.0512 | 0.6291 | 0.3467±0.0339 | 0.3885 | 0.6412±0.0304 | 0.6819 |
| DSCF-Net | 0.4746±0.0268 | 0.5025 | 0.6147±0.0249 | 0.6401 | 0.3683±0.0265 | 0.4028 | 0.7249±0.0233 | 0.7526 |
| DS$^2$CF-Net | **0.5003±0.0338** | **0.5313** | **0.6365±0.0268** | **0.6702** | **0.4012±0.0278** | **0.4388** | **0.7534±0.0202** | **0.7816** |
| | ETH80 | | USPS | | Hnd-D | | Hnd-C | |
| | Mean±std | Best | Mean±std | Best | Mean±std | Best | Mean±std | Best |
| MCF | 0.5429±0.0327 | 0.5908 | 0.5080±0.0299 | 0.5407 | 0.1695±0.0038 | 0.1746 | 0.1720±0.0102 | 0.1895 |
| GMCF | 0.5932±0.0358 | 0.6372 | 0.5238±0.0248 | 0.5523 | 0.1778±0.0043 | 0.1834 | 0.1909±0.0161 | 0.2226 |
| DSCF-Net | 0.6144±0.0250 | 0.6421 | 0.5384±0.0202 | 0.5670 | 0.2558±0.0198 | 0.2820 | 0.2442±0.0175 | 0.2700 |
| DS$^2$CF-Net | **0.6425±0.0233** | **0.6742** | **0.5566±0.0185** | **0.5833** | **0.3801±0.0317** | **0.4196** | **0.3026±0.0179** | **0.3294** |
| | Fashion MNIST | | HP2 | | TDT2 | | SCC | |
| | Mean±std | Best | Mean±std | Best | Mean±std | Best | Mean±std | Best |
| MCF | 0.5987±0.0416 | 0.6540 | 0.6759±0.0282 | 0.7112 | 0.2144±0.0367 | 0.2622 | 0.5099±0.0374 | 0.5550 |
| GMCF | 0.6181±0.0353 | 0.6563 | 0.6883±0.0241 | 0.7228 | 0.2807±0.0382 | 0.3332 | 0.5405±0.0297 | 0.5740 |
| DSCF-Net | 0.6524±0.0250 | 0.6852 | 0.7128±0.0172 | 0.7362 | 0.3258±0.0153 | 0.3423 | 0.5579±0.0244 | 0.5855 |
| DS$^2$CF-Net | **0.6771±0.0150** | **0.6985** | **0.7337±0.0173** | **0.7582** | **0.3409±0.0204** | **0.3691** | **0.5699±0.0214** | **0.5968** |

characteristic; (4) for unsupervised single-layer methods, RFA-LCF achieves the best record, since it improves the representation learning in threefold, i.e., robustness against noise, flexible reconstruction error, and enhancing the locality and sparsity by integrating the adaptive reconstruction weighting.

*Clustering with different numbers of layers for multilayer CF methods.* For the deep/multilayer algorithms, MCF, GMCF, DSCF-Net and DS²CF-Net, the number of layers varies from 1 to 10 with interval 1. And we test clustering performance in two cases, i.e., the value of K is set to 3 and the value of K is set to 6 for each database. The obtained AC and F-measure curves under the setting of K=3 over tested databases are shown in Figs.18-19 respectively. The AC and F-measure curves under the setting of K=6 over are shown in Figs. 20-21. Note that the averaged values of AC and F-measure according to the curves in Figs.18-21 are summarized in Tables VI-IX, respectively. From the results, we can find that: 1) the increase of the number of layers can generally improve the clustering results, which implies that discovering hidden deep features can indeed improve the representation learning and clustering performance. However, when the number of layers grows more than 3, the performance of each method tends to decline, which implies not the case that the more number of layers the better, since excessive decomposition may lose important information; 2) the clustering results of MCF and GMCF go down apparently as the number of layers passes 3 in most cases, which maybe because MCF and GMCF cannot ensure the intermediate representation from previous layer to be a good representation for subsequent layers. This observation result indicates that the multi-layer structures of directly feeding learnt representation from the last layer to the next layers is not reasonable; (3) DS²CF-Net performs better than all the other methods, which is mainly because it designs a novel deep coupled factorization model, and incorporates the dual structure and label constraints into CF, in addition to utilizing the partial labeled data.

## VI. CF-Based Applications

As a hot and fundamental research topic in the fields of machine learning and data mining, CF and its variants have been widely used in broad application areas. In what follows, we will summarize the application areas of the CF-based methods.

*1) Dimensionality reduction.* Real-world emerging applications usually suffer from the curse of dimensionality, so feature extraction and dimensionality reduction are fundamental problems in the areas of pattern recognition and machine learning. Note that CF and its variants have been extensively used as feature extractors for handling high-dimensional data to learn low-dimensional compact features. Specifically, for a high-dimensional $D \times N$ data matrix $X$, CF methods can learn a low-dimensional $d \times N$ representation matrix $V^T$ to represent $X$, where $d$ is the number of reduced dimension, $d<<D$. Therefore, CF can be clearly used as a preprocessing step for subsequent application tasks like data classification and clustering, which can potentially improve the performance than directly using original raw data, since the dimensionality reduction process can effectively remove redundant information and unfavorable features.

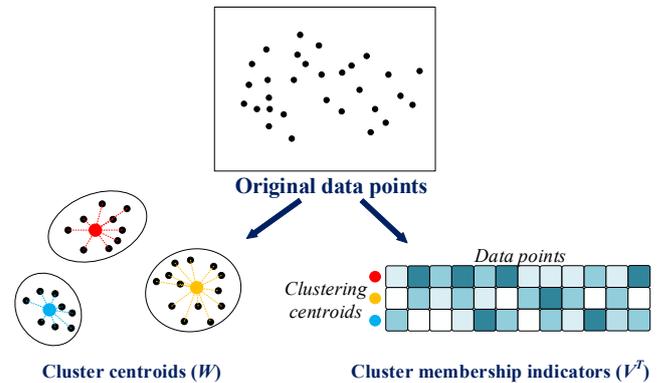

**Fig.22:** CF in data clustering task.

*2) Data clustering.* For representation learning, all CF-based methods aim at decomposing the data matrix $X$ into the product of $X$, $W$ and $V^T$, where $XW$ forms the basis vectors and $V^T$ can also be called coefficients. Note that the coefficients have a similar effect to the similarity weight matrix or adjacent matrix in graph representation learning, i.e., measuring the grouping effects of vertices. As such, we can consider the basis vectors as the cluster centroids and consider the coefficients as the indicators for cluster membership. In this way, CF and its variants can also be used as the clustering methods, which is shown in Fig. 22. However, different from the traditional clustering algorithm K-means that can be seen as a "hard" clustering method, CF-based methods can be regarded as a kind of "soft" clustering algorithm. That is, K-means algorithm clearly groups each sample into only one category, while CF and its variants can group each sample into different categories by giving the matrix $V^T$ to guide clustering since the entries of $V^T$ can be seen as the possibilities of a data point belong to a cluster. Hence, CF-based methods will be more flexible than the traditional hard clustering methods in dealing with practical applications.

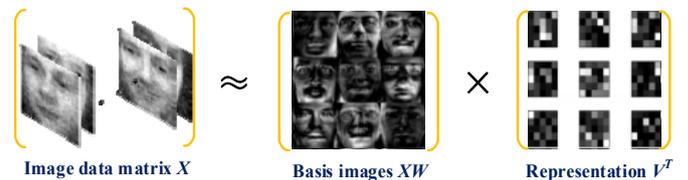

**Fig.23:** CF in image processing scenario.

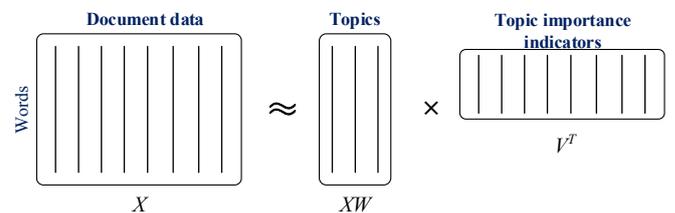

**Fig.24:** CF in text processing scenario.

*3) Image processing.* When input is image, CF-based methods can also be used for dealing with image processing tasks. Based on original image data matrix, where each column vector contains the vectorized representation of images, i.e., stacked pixel values, CF-based methods similarly decompose it into the basis vectors and representation matrix, which is exhibited in Fig.23. Due to the nonnegative constraints, the basis vectors can also be seen as the basis images containing the local parts-based information of the original images. For example, for a face image, the basis vectors can contain the nose, eye or mouth parts in the face images. Then, based on the learnt parts-based features and low-dimensional representation, CF-based algorithms can perform various image processing tasks, for instance image recognition, image reconstruction, image segmentation, image restoration and image fusion, etc.

*4) Text/document processing.* Document data processing plays an important role in the area of text data mining. Since the textual data is not only informative but also generally un-structured, the data processing tasks are usually difficult in reality. In order to obtain required useful information from a large pool of document data quickly and accurately, recent years have witnessed lots of efforts on document data processing, such as document clustering. In addition, typical document data is usually stored in the form of matrix, where each sample is always characterized by a high-dimensional and sparse vector and each element indicate the frequency of occurrence of a certain word or character. Note that the root CF algorithm was originally proposed to tackle the issue of document clustering [6], as shown in Fig.24. Specifically, for the document clustering tasks, the input is just the document data matrix. In such cases, the obtained base matrix $XW$ contains $r$ concept centers (or topics in this task). Note that CF-based methods can not only obtain the concepts, but also obtain the new representation of document data, which can be used for various processing tasks, e.g., document clustering, retrieval and categorization.

*5) Recommendation system.* Matrix factorization methods, including the CF-based methods, can also be used in the recommendation system scenario. Note that in practical situations the input data is usually a rating matrix, which represents the users' evaluations on movies, style of dress, quality of papers, review and publication speed, books, visual effects and so on. For example, users can be recommended to submit their feedback in a form of rating numbers, e.g., 1 to 5, for movies they have seen, as shown in Fig. 25. Then we can obtain two association matrices by decomposing the rating matrix using CF-based methods, where the first one is the "User-Type" matrix that indicates the users' preference for different movies and the other one is called "Type-Movie" matrix that indicates the connects of movies and the movie types. Based on jointly referring to these two matrices obtained by CF algorithms, we can then recommend movies to the users according to their favorite movie type.

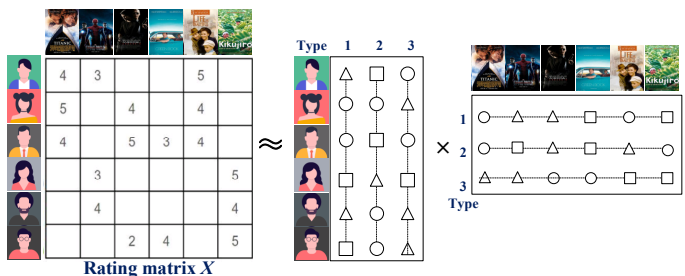

**Fig.25:** CF in recommendation system scenario.

*6) Other applications.* Due to the enhanced semantic interpretability based on the nonnegativity and ensured sparsity, CF and its extensions can also be potentially used in many other application areas, such as spectral data analysis, speech recognition, temporal segmentation, signal processing, microarray analysis, robot control, hypergraph analysis, gene sequence analysis and chemical engineering. However, these applications will not be introduced in detail due to the page limitation.

## VII. CONCLUSION AND FUTURE DIRECTIONS

### A. Conclusion Remarks

Concept factorization is a classical and popular matrix factorization technique for representation learning and data clustering, which has attracted considerable attention and broad interests in the areas of intelligent information processing and data mining. However, it is still lack of a comprehensive summarization of the existing CF-based methods. As such, this paper contributes a review to introduce the existing CF methods from different perspectives, which we hope to be able to benefit the beginners and young researchers in this field. To be specific, we differentiate current CF-based methods from two perspectives: 1) shallow vs. deep/multilayer; 2) unsupervised, fully-supervised vs. semi-supervised. The relationship and difference of methods in each category are also analyzed for better understanding. Besides the theoretical analysis, extensive experimental results are reported to compare the representation learning performance of different CF algorithms including both shallow and deep models. From the investigated cases, we find that deep models can generally deliver higher performance than corresponding shallow models, which verify that discovering hidden deep features can indeed improve the representation performance. Some possible applications of CF-based methods are also discussed.

Based on the analysis and investigation of this study, we can also conclude that CF-based methods have close relations to the graph representation learning [191-192], from the aspects of representing the nodes in graph and encoding the similarities/grouping effects. Firstly, from the framework of standard CF algorithm in Fig.2, we can conclude that: 1) when constructing the bases $U$ by $XW$, the coefficient $W$ can play a role as the graph adjacency matrix in graph representation learning to encode the local geometry structures and preserve the geometric relations within the samples of $X$; 2) from the reconstruction $XWV^T$, we can similarly conclude that both $WV^T$ and $V^T$ can be regarded as the coefficient matrix to encode the similarities/local geometry structures within the data $X$ and bases $XW$, respectively. Secondly, the columns of the new representation $V^T$ can be seen as the low-dimensional embedding, which is similar to the node embedding in graph representation learning. Note that in graph representation learning node embedding mainly aims at encoding nodes as low-dimensional vectors that summarize their graph positions and the structures of their neighborhoods. As such, by corresponding each node to the sample in data matrix $X$ for feature learning, shallow CF methods definitely have close relations to the graph representation learning in terms of edge weighting and node embedding. Thirdly, deep CF models can mine hidden hierarchical information in data, so deep CF models will be able to similarly act as the deep graph representation learning for learning deep structure information of graphs, e.g., calculating the edge weights and representing nodes to obtain the deep embedding in deep subspace.

*B. Future Work*

However, there is still a lot of space to be explored on the topic of CF-based methods in future. We simply put forward some future directions for further research on the study of CF:

(1) *Research on the optimization problem*. Existing CF-based methods suffer from one common drawback, i.e., their optimization problems can obtain a local minimum, but cannot guarantee the global optimal. Thus, it is worth studying more effective optimization methods for CF to obtain global optimal solutions, so that the numerical CF stability can be improved;

(2) *Research on the initialization of W and V*. For the moment, there is still lack of optimal solution to initialize the basis vectors $W$ and coefficient matrix $V$ for the representation learning, since they both are often initialized to be random matrices [105-113]. In other words, it is still an open issue to initialize the two nonnegative matrices $W$ and $V$ optimally. Thus, finding a more reliable approach to initialize $W$ and $V$ will have the potential to speed up the convergence of CF-based methods and more importantly improve the representation ability of features;

(3) *Research on the selection of rank r*. The rank $r$ of the factorization is undoubtedly the core parameter of each CF-based model, which directly determines the representation ability and the dimension of the resulting feature space. Similar to the number of the reduced dimension in dimensionality reduction and feature extraction, how to select an optimal value of $r$ is also an open problem that needs further investigation. In addition, exploring the theoretical guarantee and interpretable factors on the relationship between the rank $r$ and the performance of CF-based methods is also an important future work;

(4) *Research on incorporating task-driven characteristic into CF*. Since the CF-based methods have broad application areas in reality, to address the practical application problems, maybe we need to consider the task-driven application requirements, and include more targeted and useful constraints into CF-based methods, so that we can obtain satisfactory representation learning results that can meet the actual needs;

(5) *Research on more powerful CF-based deep neural networks*. It is clear that most existing CF-based methods are "shallow" models that fail to reveal the deep hidden information and hierarchical structure information from the observed input. Although certain efforts have been made to improve the representation ability by CF, the performance improvement is still not significant. Inspired by the success of deep learning and deep neural networks [165-169], to enable the CF-based methods to have a strong representation learning ability as them to well discover hidden deep features and to deal with real-world large-scale tasks, it will be useful and important to explore how to appropriately integrate the frameworks of the CF and deep neural networks, such as Convolution Neural Network (CNN), for making a breakthrough. Thus, how to design more effective deep learning architectures and structures for the deep/multi-layer CF-based methods should be investigated in future;

(6) *Research on the evaluation criteria*. Although a lot of CF-based variants have been put forward continuously in the recent decade based on different perspectives and merits, it is still lack of the systematic evaluation metrics or quantitative indicators to compare their performance in terms of a uniform standard. As such, it is also necessary to build a uniform evaluation metric to measure the performance of the CF-based methods.

ACKNOWLEDGMENT

This work is partially supported by the National Natural Science Foundation of China (61672365, 62072151), Anhui Provincial Natural Science Fund for Distinguished Young Scholars (2008085J30), and the Fundamental Research Funds for Central Universities of China (JZ2019HGPA0102). Zhao Zhang and Mingliang Xu are the corresponding authors of this paper.